\documentclass[journal]{IEEEtran}
\usepackage{amsfonts}
\usepackage{float}
\usepackage[comma,numbers,square,sort&compress]{natbib}
\usepackage{epstopdf}
\usepackage{subfigure}
\usepackage{graphicx}
\usepackage{amsthm}
\usepackage[top=2cm, bottom=2cm, left=2cm, right=2cm]{geometry}
\usepackage{algorithm}
\usepackage{algorithmicx}
\usepackage{algpseudocode}
\usepackage{amsmath}
\usepackage{multirow}
\newcommand{\RNum}[1]{\uppercase\expandafter{\romannumeral #1\relax}}

\usepackage{fancyhdr}

\begin{document}

\title{A Novel Deep Neural Network Architecture for Mars Visual Navigation}

\author{Jiang~Zhang,
        Yuanqing~Xia,
        Ganghui~Shen
\thanks{Jiang Zhang, Yuanqing Xia and Ganghui Shen are with the School of Automation, Beijing Institute of Technology, Beijing 100081, China. Email: bitzj2015@outlook.com (Zhang), xia\_yuanqing@bit.edu.cn (Xia), hxyzsgh@gmail.com (Shen).}}

\maketitle
\thispagestyle{fancy}
\fancyhead{} 
\lhead{} 
\lfoot{\begin{small}
~\copyright~2018 IEEE. Personal use of this material is permitted. Permission from IEEE must be obtained for all other uses, in any current or future media, including reprinting/republishing this material for advertising or promotional purposes, creating new collective works, for resale or redistribution to servers or lists, or reuse of any copyrighted component of this work in other works.
\end{small}}
\cfoot{} 
\rfoot{}

\begin{abstract}
In this paper, emerging deep learning techniques are leveraged to deal with Mars visual navigation problem. Specifically, to achieve precise landing and autonomous navigation, a novel deep neural network architecture with double branches and non-recurrent structure is designed, which can represent both global and local deep features of Martian environment images effectively. By employing this architecture, Mars rover can determine the optimal navigation policy to the target point directly from original Martian environment images. Moreover, compared with the existing state-of-the-art algorithm, the training time is reduced by 45.8\%. Finally, experiment results demonstrate that the proposed deep neural network architecture achieves better performance and faster convergence than the existing ones and generalizes well to unknown environment.
\end{abstract}


\section{Introduction}

\begin{figure}[H]
\centering
\includegraphics[width=0.75\linewidth]{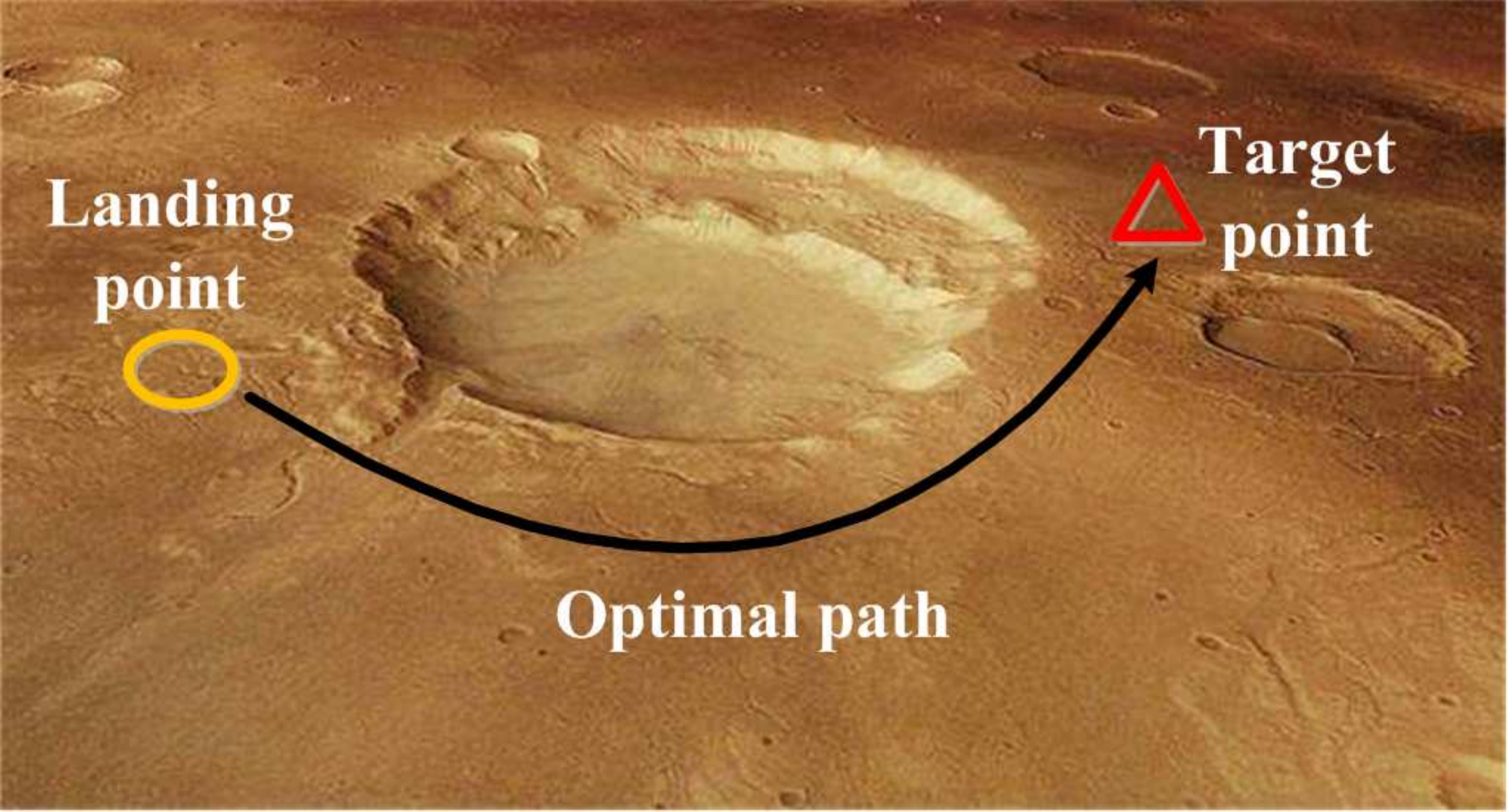}
\caption{Visual navigation phase after landing.}\label{Fig.1.1}
\end{figure}

Generally, there are three basic phases during Mars exploration missions---entry, descent and landing (EDL) \cite{ref1}, where the landing phase finally determines whether Mars rovers land on Martian surface safely and precisely. Due to the large uncertainties and dispersions derived from Martian environment, existing algorithms used for EDL phase cannot guarantee the precision of the  Mars rovers' landing on the target point. Moreover, after landing, Mars rovers are usually required to move to new target points constantly in order to carry out new exploration tasks. Hence, in future Mars missions, autonomous navigation algorithms are essential for Mars rovers to avoid risky areas (such as craters, high mountains and rocks) and reach target points precisely and efficiently (Fig.~\ref{Fig.1.1}).

Currently, one of the most significant methods for Mars navigation is visual navigation \cite{ref2}. Two main methods for Mars visual navigation are blind drive and autonomous navigation with hazard avoidance (AutoNav) \cite{ref3}. In blind drive, all commands for Mars rovers are determined by engineers from the earth before starting missions. This method promptly reduces the efficiency and flexibility of exploration missions. By contrast, AutoNav can lender Mars rover execute missions unmannedly. Thus, it is more in coincidence with the increasing future demand on Mars rovers' autonomy and intelligence.

Classical algorithms for AutoNav such as \emph{Dijkstra} \cite{ref4,ref5}, $A^*$ \cite{ref6,ref7} and $D^*$ \cite{ref8,ref9} have been widely researched in the past decades. It is noteworthy that these algorithms have to search the optimal path iteratively on cellular grip maps, which are both time consuming and memory consuming \cite{ref10}. When dimensions of maps become large and computation resources are limited, these algorithms may fail to offer the optimal navigation policy. To overcome the dimension explosion problem, intelligent algorithms such as neural network \cite{ref11}, genetic algorithm \cite{ref12} and particle swarm algorithm \cite{ref13} were extended into planetary navigation problem. However, prior knowledge about the obstacles in maps is prerequisite for these algorithms to work. 

To provide the optimal navigation policy directly from natural Martian scenes, effective feature representation algorithms are required. That is, these algorithms have to understand deep features of input image such as the shape and location of obstacles firstly and then determine the navigation policy according to these deep features. In recent years, \emph{Deep Convolutional Neural Networks (DCNNs)} have received wide attention in computer vision field for their superior feature representation capability \cite{ref14}.
Notably, although the training process of \emph{DCNNs} consumes massive time and computation resource, it is completed offline. When applying \emph{DCNNs} to represent deep features of images online after training, it costs little time and computation resource. Therefore, \emph{DCNNs} have been widely applied in varieties of visual tasks such as image classification \cite{ref15}, object detection \cite{ref16}, visual navigation \cite{ref17} and robotic manipulation \cite{ref18}.

Inspired by the state-of-art performance of \emph{DCNNs} in computer vision field, planetary visual navigation algorithms based on deep neural network have been researched. In \cite{ref19}, a 3 dimensional \emph{DCNN} was designed to create a safety map for autonomous landing zone detection from terrain image. In \cite{ref20}, a \emph{DCNN} was trained to predict rover's position from terrain images for Lunar navigation. Though these algorithms are capable of extracting deep features of raw images, they are unable to provide the optimal policy for navigation directly. To solve this probelm, \emph{Value Iteration Network (VIN)} was firstly proposed in \cite{ref21} to plan path directly from images and applied to Mars visual navigation problem successfully. Then, in \cite{ref22}, \emph{Memory Augmented Control Network} was proposed to find the optimal path for rovers in partially observable environment. Both of these networks for visual navigation employed \emph{Value Iteration Module}. However, it takes massive time to train them.

In this paper, an efficient algorithm to determine the optimal navigation policy directly from original Martian images is investigated. Specifically, a novel \emph{DCNN} architecture with double branches is designed. It can represent both global and local deep features of input images and then achieve precise navigation efficiently. The main contributions of this paper are summarized as follows:
\begin{itemize}
\item Emerging deep learning techniques (deep neural networks) are leveraged to deal with Mars visual navigation problem.
\item The proposed \emph{DCNN} architecture with double branches and non-recurrent structure can find the optimal path to target point directly from global Martian environment images and prior knowledge about risky areas in images are not required.
\item Compared with acknowledged (\emph{VIN}), the proposed \emph{DCNN} architecture achieves better performance on Mars visual navigation and the training time is reduced by 45.8\%.
\item The accuracy and efficiency of this novel architecture are demonstrated through experiment results and analysis.
\end{itemize}

The rest paper is organized as follows. Section \RNum{2} provides preliminaries of this paper. Section \RNum{3} describes the novel \emph{DCNN} architecture for Mars visual navigation. Experimental results and analysis are illustrated in Section \RNum{4}, followed by discussion and conclusions in Section \RNum{5}.
\section{Preliminaries}
\subsection{Markov Decision Process}
\begin{figure}[H]
\centering
\includegraphics[width=1\linewidth]{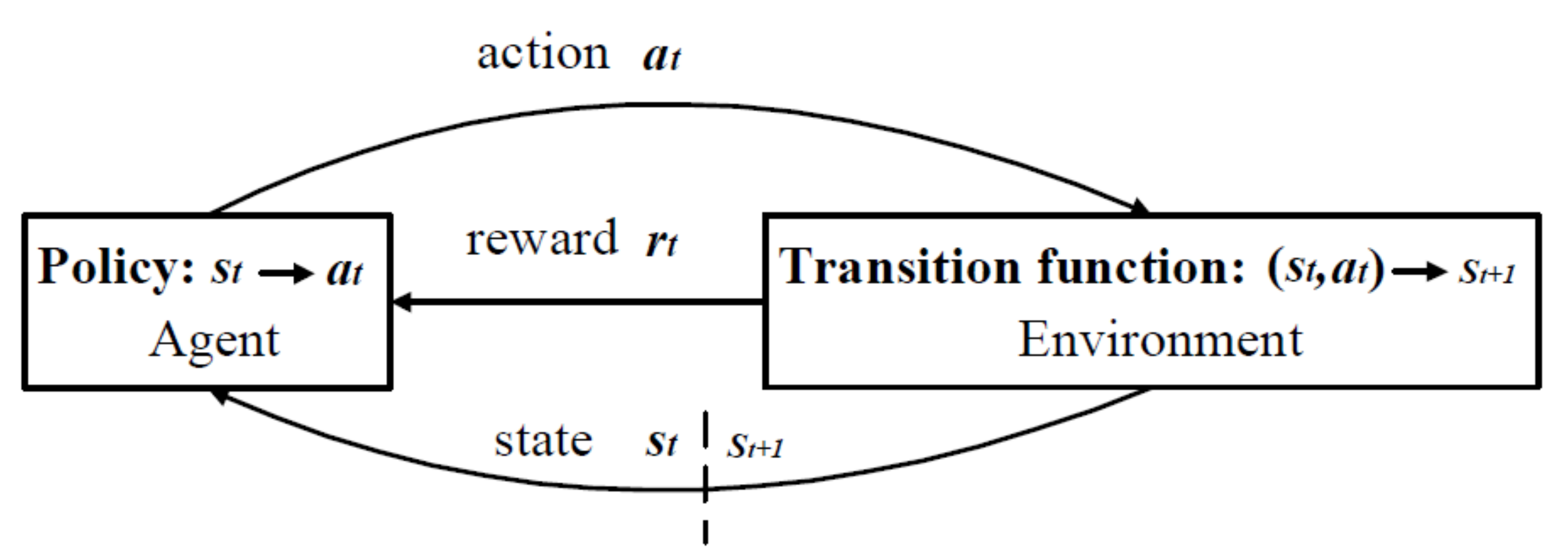}
\caption{Markov decision process.}\label{Fig.2.1}
\end{figure}

Mars visual navigation can be formulated as a \emph{Markov Decision Process (MDP)}, since the next state of Mars rover can be determined by its current state and action completely. A standard \emph{MDP} for sequential decision making is composed of action space $\mathcal{A}$, state space $\mathcal{S}$, reward $r:\mathcal{S}\times \mathcal{A}\rightarrow \mathcal{R}$, transition probability distribution $P:\mathcal{S}\times \mathcal{A} \times \mathcal{S}\rightarrow \mathcal{R}$ and policy $\pi_{\theta}: \mathcal{S}\times \mathcal{A}\rightarrow \mathcal{R}$. At time step $t$, the agent can obtain its state $s_t \in \mathcal{S}$ from environment and then choose its action $a_t$ satisfying distribution $a_t \sim \pi_{\theta}(a|s_t), a \in \mathcal{A}$. After that, its state will transit into $s_{t+1}\in \mathcal{S}$ and the agent will then receive reward $r_t=r(s_t,a_t) \in \mathcal{R}$ from environment, where $s_{t+1} \in \mathcal{S}$ satisfies the transition probability distribution $s_{t+1}\sim P(s|s_t,a_t), s \in \mathcal{S}$. The whole process is shown in Fig.~\ref{Fig.2.1}.

Denote the discount factor of reward by $\gamma \in \mathcal{R}$. A policy is defined as optimal if and only if its parameter $\theta$ satisfies
\begin{equation}
\label{eq2.1}
\theta^{*} = \arg\max_{\theta} E_{s_0,a_0,s_1,a_1,...}[r_0+\gamma r_1+\gamma^{2} r_2+...],
\end{equation}

To measure the expected accumulative reward of $s_t$ and $(s_t,a_t)$, the state value function and the action value function are defined respectively as 
\begin{equation}
\label{eq2.2}
V^{\pi_{\theta}}(s_t)=E_{a_t,s_{t+1},a_{t+1}...}[\sum_{i=t}^{+\infty}\gamma^{i-t}r(s_{i},a_{i})],
\end{equation}
\begin{equation}
\label{eq2.3}
\begin{aligned}
Q^{\pi_{\theta}}(s_t,a_t)&=r_{t}+E_{s_{t+1},a_{t+1}...}[\sum_{i=t+1}^{+\infty}\gamma^{i-t}r(s_{i},a_{i})]\\
&=r_{t} + E_{s_{t+1}}[V^{\pi_{\theta}}(s_{t+1})].
\end{aligned}
\end{equation}

By substituting Eq.~(\ref{eq2.2}) into Eq.~(\ref{eq2.1}), the following equation is derived as
\begin{equation}
\label{eq2.4}
\theta^{*} = \arg\max_{\theta} E_{s_0}[V^{\pi_{\theta}}(s_0)].
\end{equation}

By solving Eq.~(\ref{eq2.4}), the optimal policy is determined such that the objective of \emph{MDP} is achieved. However, since both state value function and action value function are unknown before, Eq.~(\ref{eq2.4}) cannot be solved directly. Therefore, state value function and action value function have to be estimated in order to solving \emph{MDP} problem.

\subsection{Value Function Estimation}
Value iteration is an typical method for value function estimation and then addressing \emph{MDP} problem \cite{ref25}. Denote the estimated state value function at step $k$ by $V_{k}(s)$, and the estimated action value function for each state at step $k$ by $Q_{k}(s,a)$. $\pi_{k}$ is utilized to represent the policy at step $k$. Then, the value iteration process can be expressed as 
\begin{equation}
\label{eq2.5}
\pi_{k}(s_i) = \arg \max_{a_i}Q_{k}(s_i,a_i) \quad (i=0,1,\cdots),
\end{equation}
\begin{equation}
\begin{aligned}
\label{eq2.6}
V_{k+1}(s_i) &= Q_{k+1}(s_i,\pi_{k}(s_i)) \\
&= r_i +  E_{s_{i+1}}[V_{k}(s_{i+1})] \quad (i=0,1,\cdots).
\end{aligned}
\end{equation}

Through iteration, the policy and value functions will converge to optimum $\pi^{*}$, $Q^{*}$ and $V^{*}$ simultaneously.

\begin{figure}[H]
\centering
\includegraphics[width=1\linewidth]{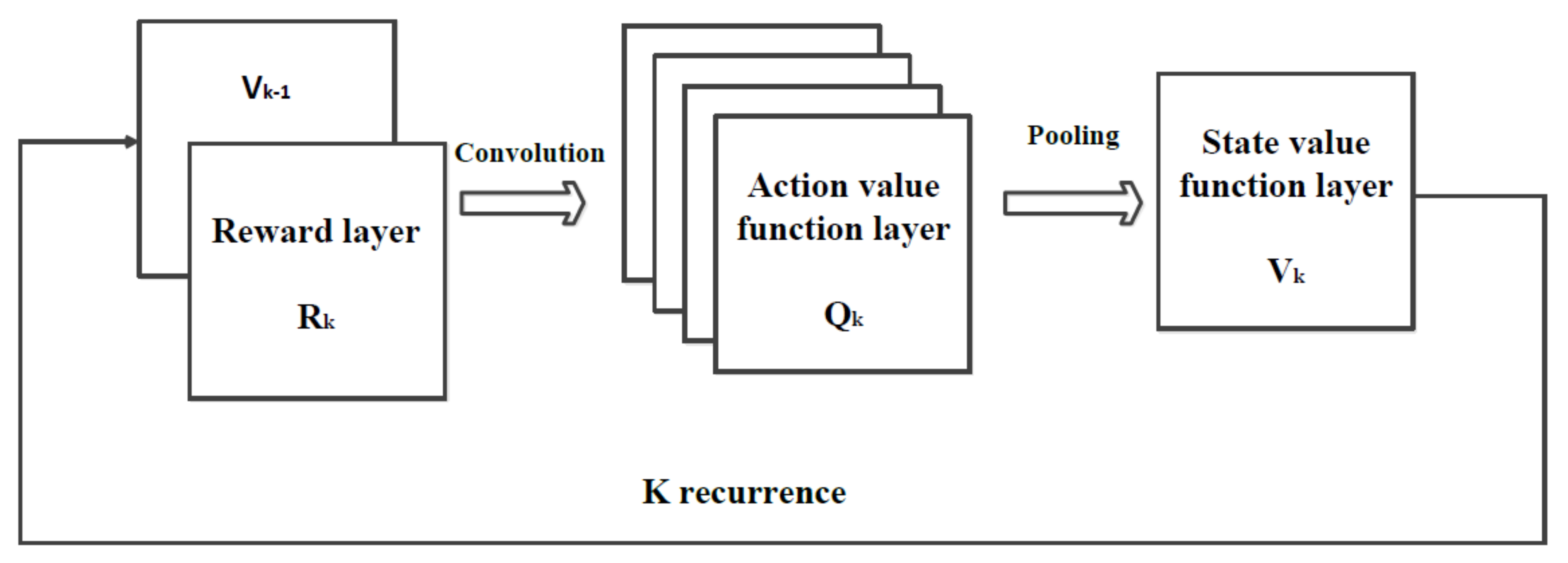}
\caption{Value iteration module.}\label{Fig2.2}
\end{figure}

However, since it is difficult to determine the explicit representation of $\pi_{k}$, $Q_{k}$ and $V_{k}$ (especially when the dimension of $s_t$ is high), \emph{VIN} is applied to approximate this process successfully. Specifically, \emph{VIN} is designed with \emph{Value Iteration Module}, which consists of recurrent convolutional layers \cite{ref21}. As illustrated in Fig.~\ref{Fig2.2}, the value function layer $V_k$ is stacked with the reward layer $R_k$ and then filtered by a convolutional layer and a max-pooling layer recurrently. Furthermore, through \emph{VIN}, navigation information including global environment and target point can be conveyed to each state in the final value function layer. Experiments demonstrate that this architecture performs well in navigation tasks. However, it takes lots of time and computation resource to train such a recurrent convolutional neural network when the value of $K$ becomes large. Therefore, replacing \emph{Value Iteration Module} with a more efficient and non-recurrent architecture without losing its excellent navigation performance becomes the focus of this paper.

\subsection{Learning-Based Algorithms}
Typically, there exist two learning-based algorithms for training \emph{DCNNs} in value function estimation---\emph{Reinforcement learning} \cite{ref25} and \emph{Imitation learning} \cite{ref26}. In \emph{Reinforcement learning}, no prior knowledge is required and the agent can find the optimal policy in complex environment by trial and error \cite{ref27}. However, the training process of \emph{Reinforcement learning} is computationally inefficient. In \emph{Imitation learning}, when the expert dataset is given $\{(s_i, y_i)\}_{i=1}^{i=N}$, the training process transforms into supervised learning with higher data-efficiency and fitting accuracy.

Considering that expert dataset $\{(s_i, y_i)\}_{i=1}^{i=N}$ for global visual navigation is available ($y_i \in \{0,1,\cdots,N\}$ is the optimal action at state $s_t$ and $N$ is the number of samples), in this paper, \emph{Imitation learning} method is applied to find the optimal navigation policy.
\section{Model Description}
\subsection{Mars Visual Navigation Model}
In this subsection, the process of formulating Mars visual navigation into \emph{MDP} is presented. More precisely, state $s_t =\{I_t,(g_{1t},g_{2t}),(x_{1t},x_{2t})\}$ is composed of the Martian environment image $I_t\in\mathcal{R}^{M\times M\times 3}$, target point $(g_{1t},g_{2t}) \in \mathcal{Z}\times\mathcal{Z}$ and the current position of Mars rover $(x_{1t},x_{2t})\in \{0,1,\cdots,M-1\}\times \{0,1,\cdots,M-1\}$ at time step $t$. The action $a_t \in \{0,1,\cdots,7\}$ represents the moving direction of the Mars rover at time step $t$ (0:east, 1:south, 2:west, 3:north, 4:southeast, 5:northeast, 6:southwest, 7:northwest). After taking action $a_t$, the current location of the Mars rover will change and the state $s_t$ will transit into $s_{t+1}$. If the Mars rover reaches the target point precisely at time $t+1$, a positive reward will be obtained (such as $r_t=1.0$). Otherwise, the Mars rover will get a negative reward (such as $r_t=-1.0$).


Furthermore, the output vector of the proposed \emph{DCNN} is defined as $\Pi_{\theta}(s_t)=[Q^{\pi_{\theta}}(s_t,a=0), \cdots, Q^{\pi_{\theta}}(s_t,a=7)]^{T}$ ($s_t \in \mathcal{S}$). Then the training loss is defined in cross entropy form with $L_2$ norm as \cite{ref28}
\begin{equation}
\begin{aligned}
\label{eq2.9}
L(\theta) = -\frac{1}{N}\sum_{i=1}^{N}Y_i log( \Pi_{\theta}(s_i)) + \lambda ||\theta||_2,
\end{aligned}
\end{equation}
where $N$ is the number of training samples, $Y_i$ is the one-hot vector \cite{ref29} of $y_i$ and $\lambda$ is the hyperparameter adjusting the effect of $L_2$ norm on the loss function.

By minimizing the loss function $L(\theta)$, the optimal parameter of navigation policy is determined as follows
\begin{equation}
\begin{aligned}
\label{eq2.10}
\theta^* = \arg \min_{\theta}L(\theta).
\end{aligned}
\end{equation}

\subsection{The Novel Deep Neural Network Architecture}
In this subsection, the novel deep neural network architecture---\emph{DB-Net} with double branches for deep feature representations and value function estimation is illuminated. The principle design idea of \emph{DB-Net} is to replace \emph{Value Iteration Module} of \emph{VIN} with a non-recurrent convolutional network structure. Firstly, the reprocessing layers of \emph{DB-Net} compresses the input Martian environment image $I_t$ into feature map $I_{t}^{'}(I_t,g_{1t},g_{2t}) \in \mathcal{R}^{N\times N\times A}$($M = lN, l\in \mathcal{Z}, A \in \mathcal{Z}$ and $N\in \mathcal{Z}$). Then, the global deep feature $f_1(I_t^{'}) \in \mathcal{R}^{B}$ ($B\in \mathcal{Z}$) and the local deep feature $f_2(I_{t}^{'},x_{1t},x_{2t}) \in \mathcal{R}^{C}$  ($C\in \mathcal{Z}$) are extracted from feature map $I_{t}^{'}$ by branch one and branch two respectively. By fusing $f_1(I_{t}^{'})$ and $f_2(I_{t}^{'},x_{1t},x_{2t})$, the final deep feature (value function estimation) $\Pi_{\theta}(s_t)=\Pi_{\theta}(I_t,g_{1t},g_{2t},x_{1t},x_{2t}) \in \mathcal{R}^{8}$ of Martian environment image is derived. Then, the optimal navigation policy can be determined through Eq.~(\ref{eq2.5}).

\begin{figure*}
\centering
\includegraphics[width=1\linewidth]{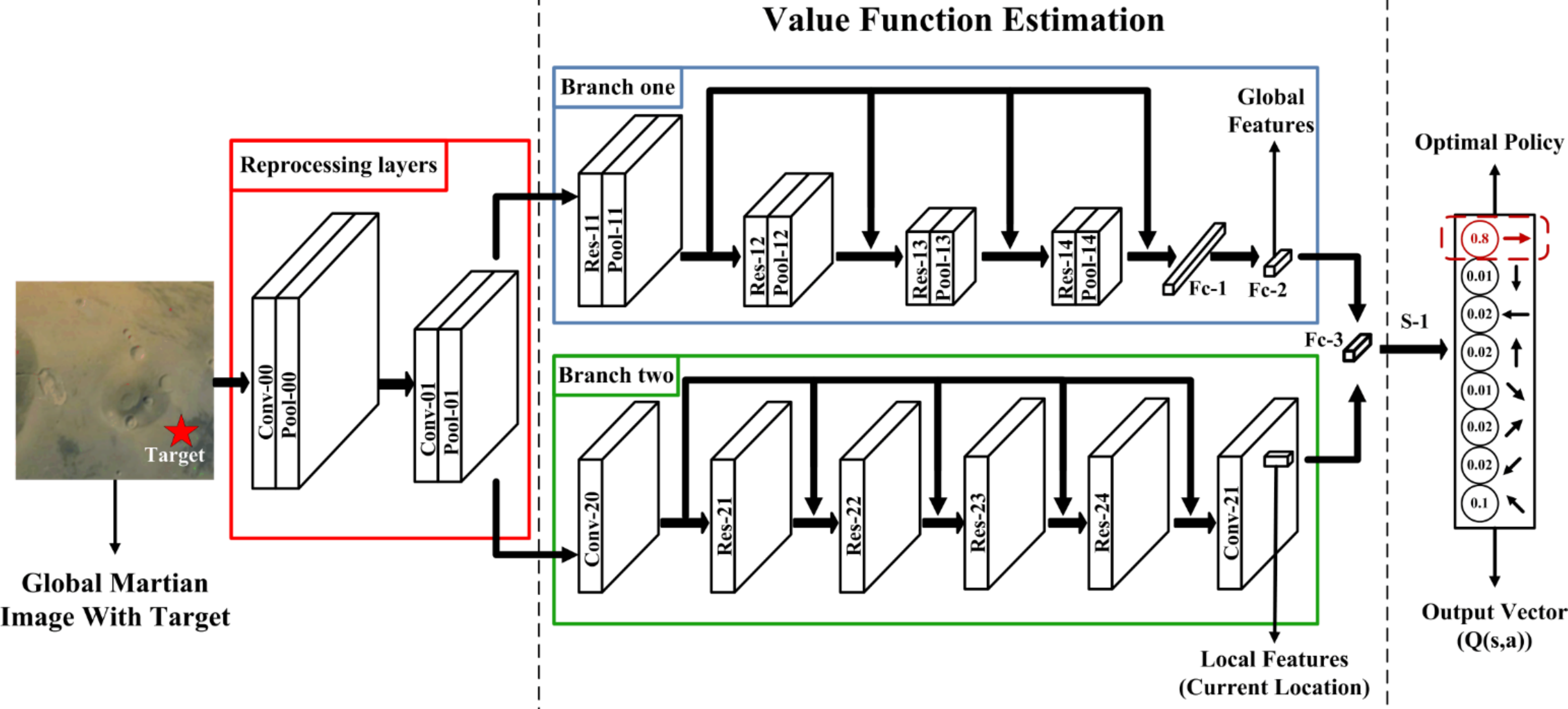}
\caption{The diagram of DB-Net.}\label{Fig.3.2}
\end{figure*}

The diagram of \emph{DB-Net} is illustrated in Fig.~\ref{Fig.3.2}, where \textbf{Conv}, \textbf{Pool}, \textbf{Res}, \textbf{Fc} and \textbf{S} are short for convolutional layer, max-pooling layer, residual convolution layer, fully-connected layer and softmax layer respectively. More specific explanations of \emph{DB-Net} are given as follows.

(1) The reprocessing layers comprises of two convolutional layers (Conv-00, Conv-01) and two max-pooling layers (Pool-00, Pool-01). After compressing the original image $I_t$, the navigation policy becomes \emph{area by area} instead of \emph{point by point} (each area has size $l\times l$). Thus, the efficiency of visual navigation is promptly enhanced.

\begin{figure}[H]
\centering
\includegraphics[width=1\linewidth]{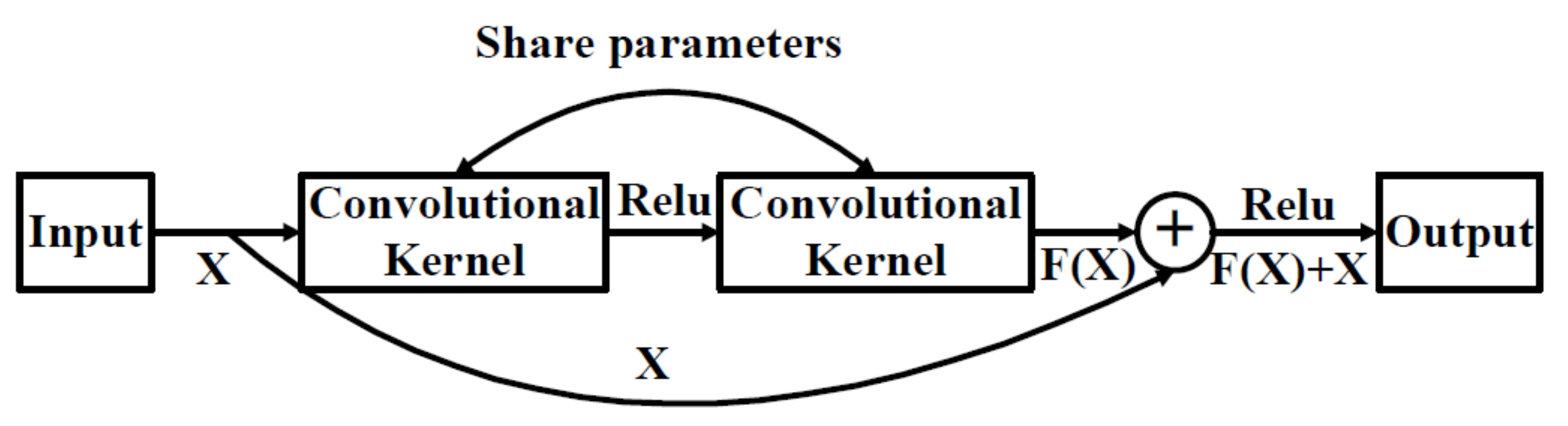}
\caption{Residual convolutional layer}\label{Fig.3.3}
\end{figure}

(2) Branch one consists of one convolutional layer (Conv-10), three residual convolutional layers (Res-11, Res-12, Res-13), four max-pooling layers (Pool-10, Pool-11, Pool-12, Pool-13) and two fully connected layers (Fc-1, Fc-2). Notably, residual convolutional layer (Fig.~\ref{Fig.3.3}) is one kind of convolutional layer proposed in \cite{ref30}, which not only increases the training accuracy of convolutional neural networks with deep feature representations, but also makes them generalize well to testing data. Considering that \emph{DB-Net} is required to represent deep features of Martian image and achieves high-precision in unknown Martian environment images, residual convolutional layers are employed on \emph{DB-Net}. The deep feature $f_1(I_t^{'})$ represented by this branch is a global guidance to the Mars rover, containing abstract information about global Martian environment $I_t$ and target point $(g_{1t},g_{2t})$.

(3) Branch two is composed of two convolutional layers (Conv-20, Conv-21) and four residual convolutional layers (Res-21, Res-22, Res-23, Res-24). The deep feature $f_2(I_t^{'},x_{1t},x_{2t})$ represented by this branch depicts the local value distribution of Martian environment image $I_t$ with target $(g_{1t},g_{2t})$, which acts as a local guidance to Mars rover.
(4) The final deep feature $\Pi_{\theta}(I_t,g_{1t},g_{2t},x_{1t},x_{2t})=\Pi_{\theta}(s_t)$ is fully connected with $f_1$ and $f_2$ through Fc-3, corresponding to the $Q^{\pi_{\theta}}$ value of one action $a_t$ at current state $s_t$. Hence, following Eq.~(\ref{eq2.5}), the optimal visual navigation policy is determined.

Compared with \emph{VIN}, not only the depth of \emph{DB-Net} is reduced significantly (since it is non-recurrent), but also both global and local information of the image is kept and represented effectively. Detailed parameters of \emph{DB-Net} are demonstrated in TABLE~\ref{tab3.1}.

\begin{table}[H]
\centering
\caption{Detailed parameters of \emph{DB-Net}}\label{tab3.1}
\begin{tabular}{|c|c|c|}
\hline
\multirow{3}{*}{Reprocessing layers}&Conv-00&$6\times5\times5$ kernels with stride 1 \\ \cline{2-3}
~&Pool-00&$3\times3$ kernels with stride 2 \\ \cline{2-3}
(A=12)&Conv-01&$12\times4\times4$ kernels with stride 1 \\ \cline{2-3}
~&Pool-01&$3\times3$ kernels with stride 2 \\
\hline
\multirow{9}{*}{Branch one}&Conv-10&$20\times5\times5$ kernels with stride 1 \\ \cline{2-3}
~&Pool-10&$3\times3$ kernels with stride 1 \\ \cline{2-3}
~&Res-11&$20\times3\times3$ kernels with stride 1  \\ \cline{2-3}
~&Pool-11&$3\times3$ kernels with stride 2 \\ \cline{2-3}
~&Res-12&$20\times3\times3$ kernels with stride 1  \\ \cline{2-3}
~&Pool-12&$3\times3$ kernels with stride 2 \\ \cline{2-3}
(B=10)&Res-13&$20\times3\times3$ kernels with stride 1  \\ \cline{2-3}
~&Pool-13&$3\times3$ kernels with stride 1 \\ \cline{2-3}
~&Fc-1&192 nodes  \\ \cline{2-3}
~&Fc-2&10 nodes \\
\hline
\multirow{6}{*}{Branch two}&Conv-20&$20\times5\times5$ kernels with stride 1 \\ \cline{2-3}
~&Res-21&$20\times3\times3$ kernels with stride 1  \\ \cline{2-3}
~&Res-22&$20\times3\times3$ kernels with stride 1  \\ \cline{2-3}
~&Res-23&$20\times3\times3$ kernels with stride 1  \\ \cline{2-3}
(C=10)&Res-24&$20\times3\times3$ kernels with stride 1  \\ \cline{2-3}
~&Res-25&$20\times3\times3$ kernels with stride 1  \\ \cline{2-3}
~&Conv-21&$10\times3\times3$ kernels with stride 1 \\
\hline
\multirow{2}{*}{Output layers}&Fc-3&8 nodes \\ \cline{2-3}
~&S-1&8 nodes \\ \cline{2-3}
\hline
\end{tabular}
\end{table}

\section{Experiments and Analysis}
In this section, \emph{DB-Net} and \emph{VIN} are firstly trained and tested on Martian image dataset derived from HiRISE \cite{ref31}. The dataset consists of 10000 high-resolution Martian images, each of which has 7 optimal trajectory samples (generated randomly). The training set and the testing set consist of 6/7 and 1/7 dataset respectively. Then, navigation accuracy and training efficiency of \emph{DB-Net} and \emph{VIN} are compared. Finally, detailed analysis of \emph{DB-Net} is made through model ablation experiments. More precisely, the following questions will be investigated:
\begin{itemize}
\item Could \emph{DB-Net} provide the optimal navigation policy directly from original Martian environment images?
\item Could \emph{DB-Net} outperform the best framework---\emph{VIN} in accuracy and efficiency?
\item Could \emph{DB-Net} keep its performance after ablating some of its components?
\end{itemize}

\subsection{Experiment Results on Martian Images}
\begin{figure}[H]
\centering
\subfigure[Channel1]{
\includegraphics[width=0.31\linewidth]{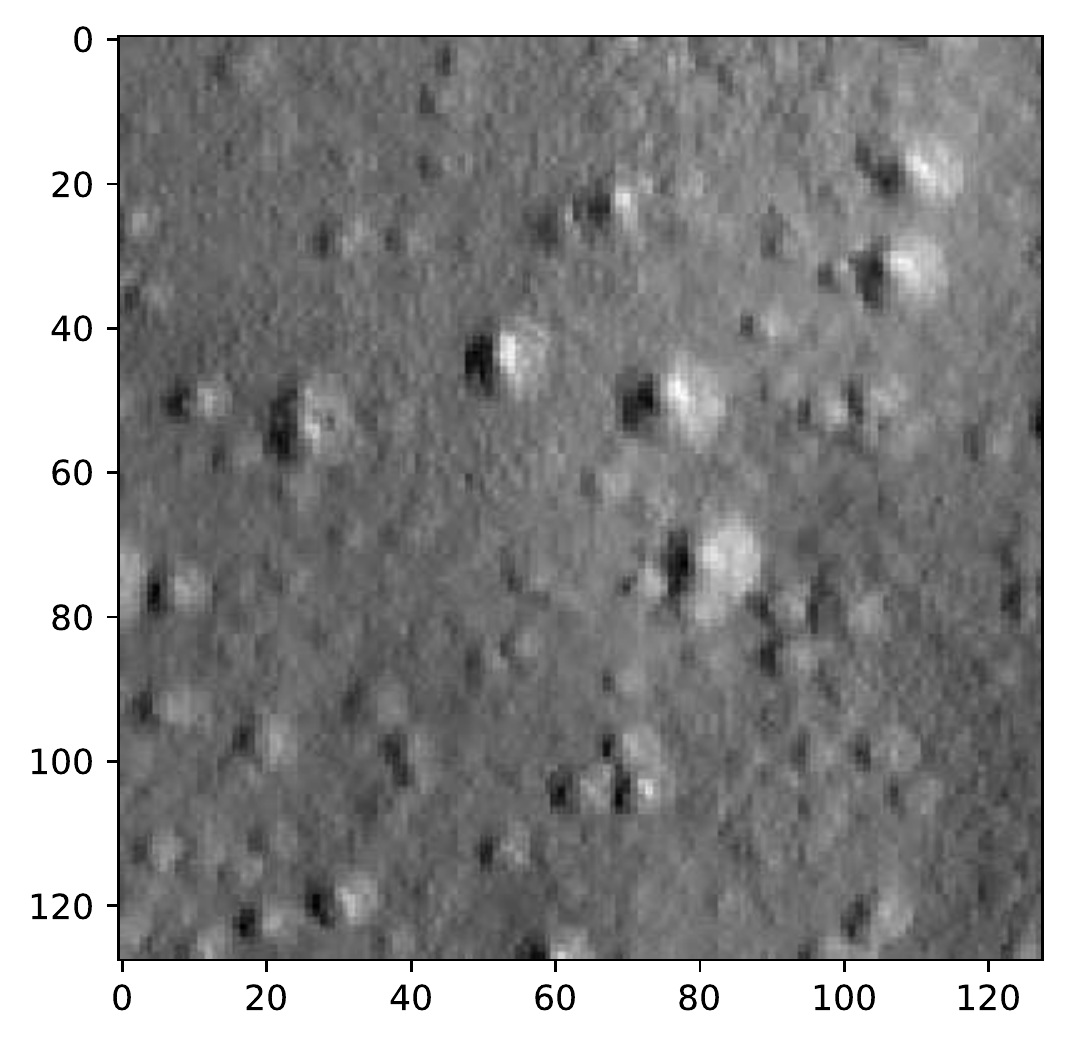}}
\subfigure[Channel2]{
\includegraphics[width=0.31\linewidth]{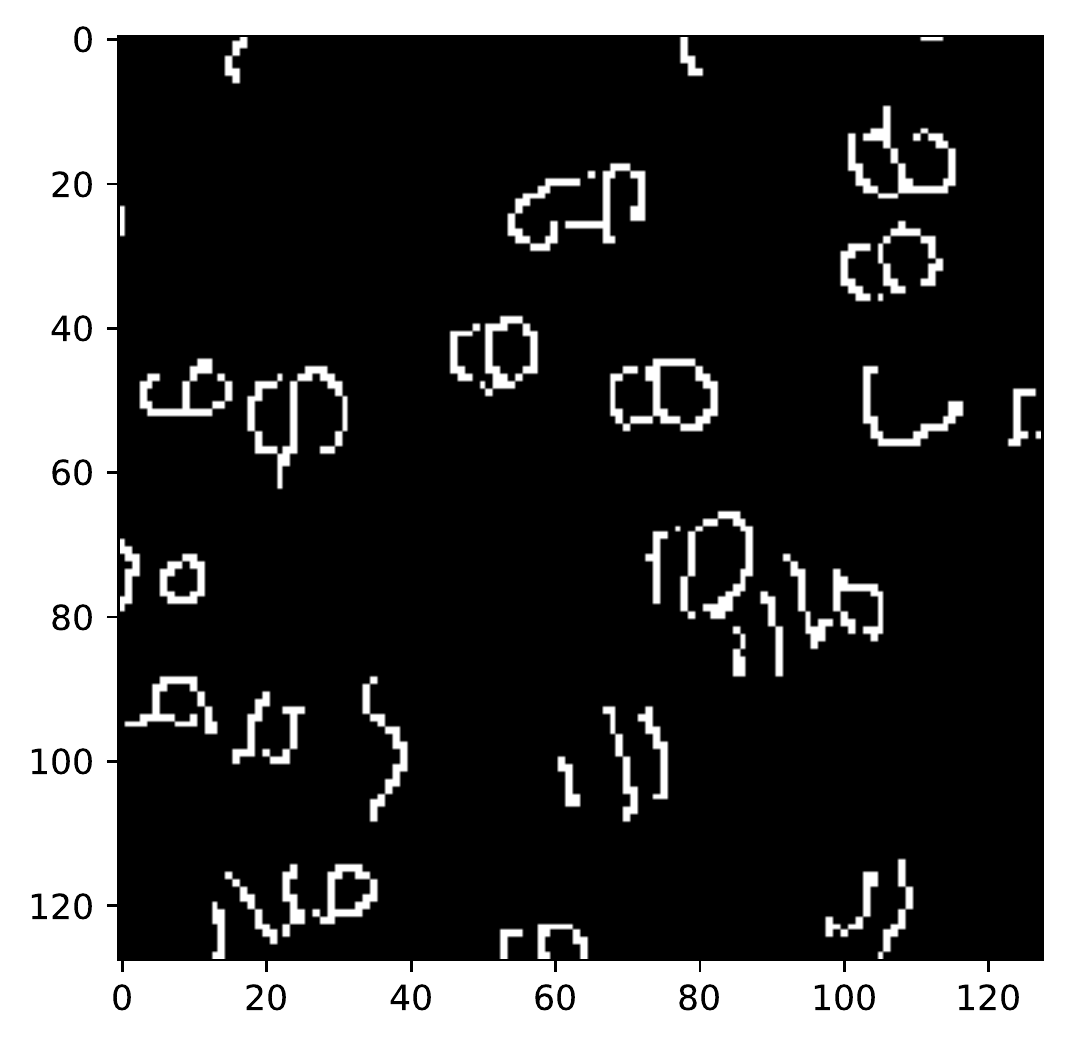}}
\subfigure[Channel3]{
\includegraphics[width=0.31\linewidth]{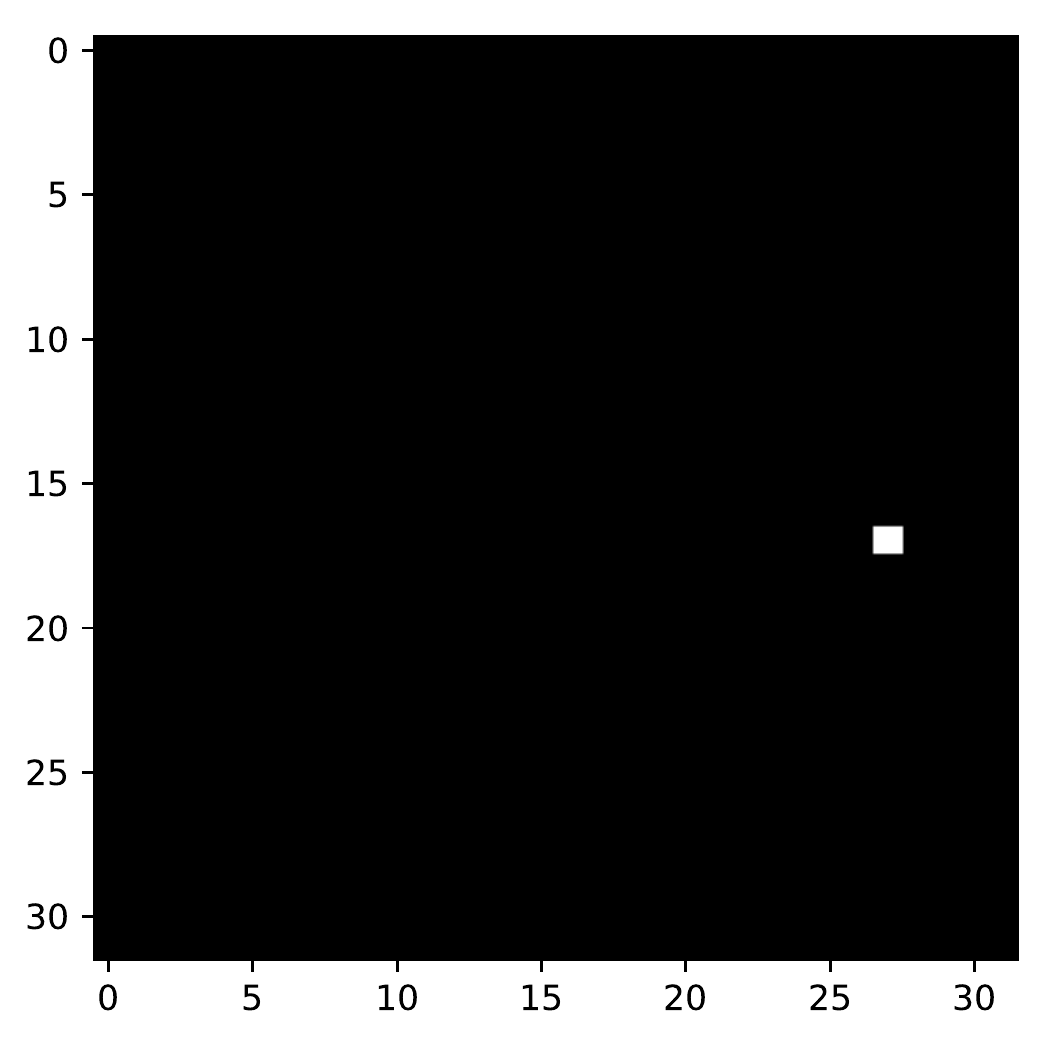}}
\caption{The input image of \emph{DB-Net}. (Channel1 is the gray image. Channel2 is the edge image. Channel3 is the target image.) }\label{Fig.4.0}
\end{figure}

In this subsection, the process of training and testing \emph{DB-Net} and \emph{VIN} on Martian image dataset is described. The input image has a size of $128\times128$ with 3 channels (i.g. $M=128$), consisting of the gray image of original Martian environment, the edge image of original Martian environment generated by \emph{Canny} algorithm \cite{ref32} and the target image (Fig.~\ref{Fig.4.0}). Then, training accuracy and testing accuracy of \emph{DB-Net} and \emph{VIN} are counted to contrast the proportion of the optimal action they take each step. To compare the navigation performance of \emph{DB-Net} and \emph{VIN}, success rate on both training images and testing images are counted. It is worth noting that a navigation process is considered successful if and only if the Mars rover reaches target point from start point without running into any risky areas.

\begin{figure}[H]
\centering
\subfigure[Training loss]{
\includegraphics[width=0.47\linewidth]{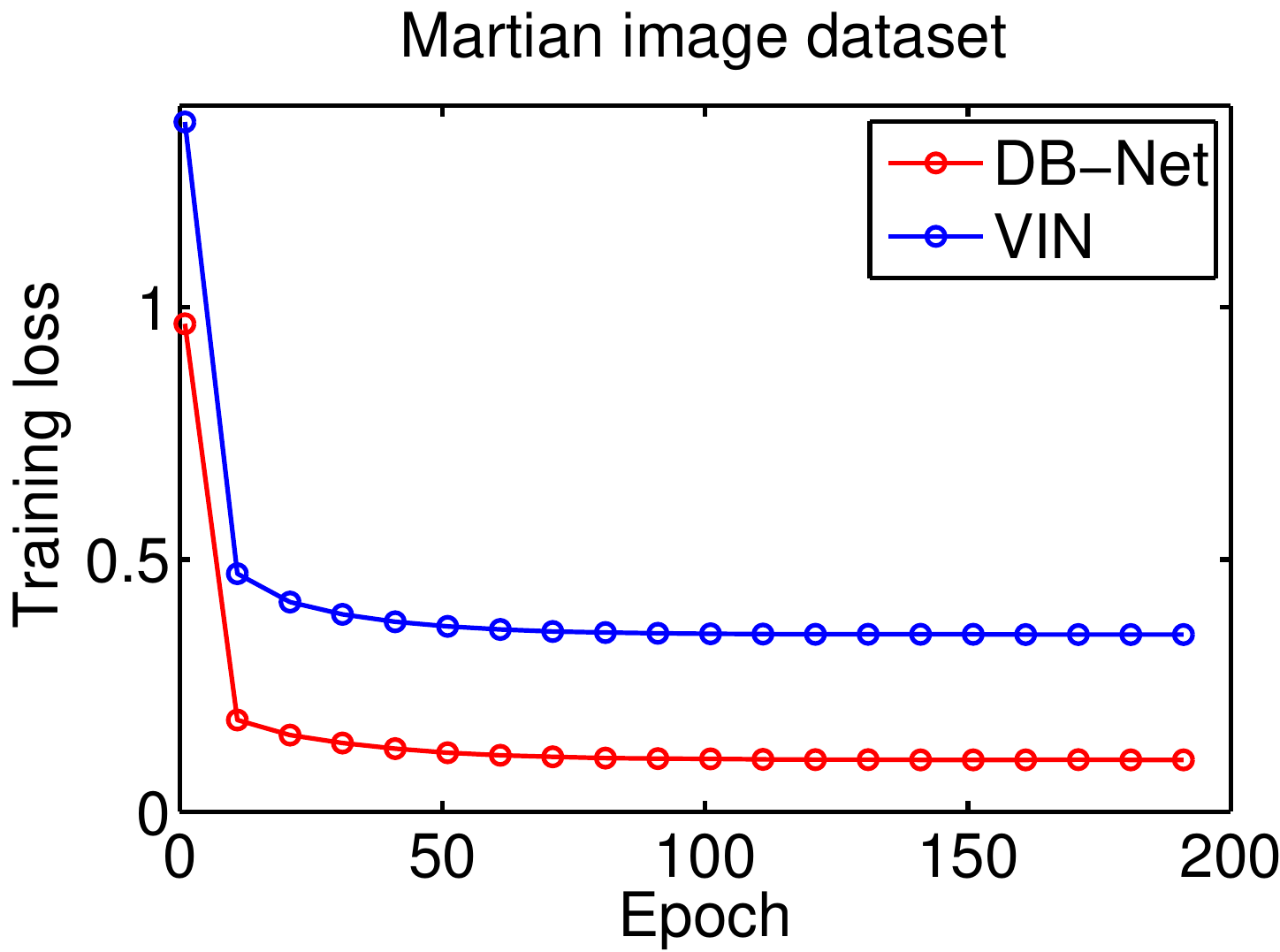}}
\subfigure[Training error]{
\includegraphics[width=0.47\linewidth]{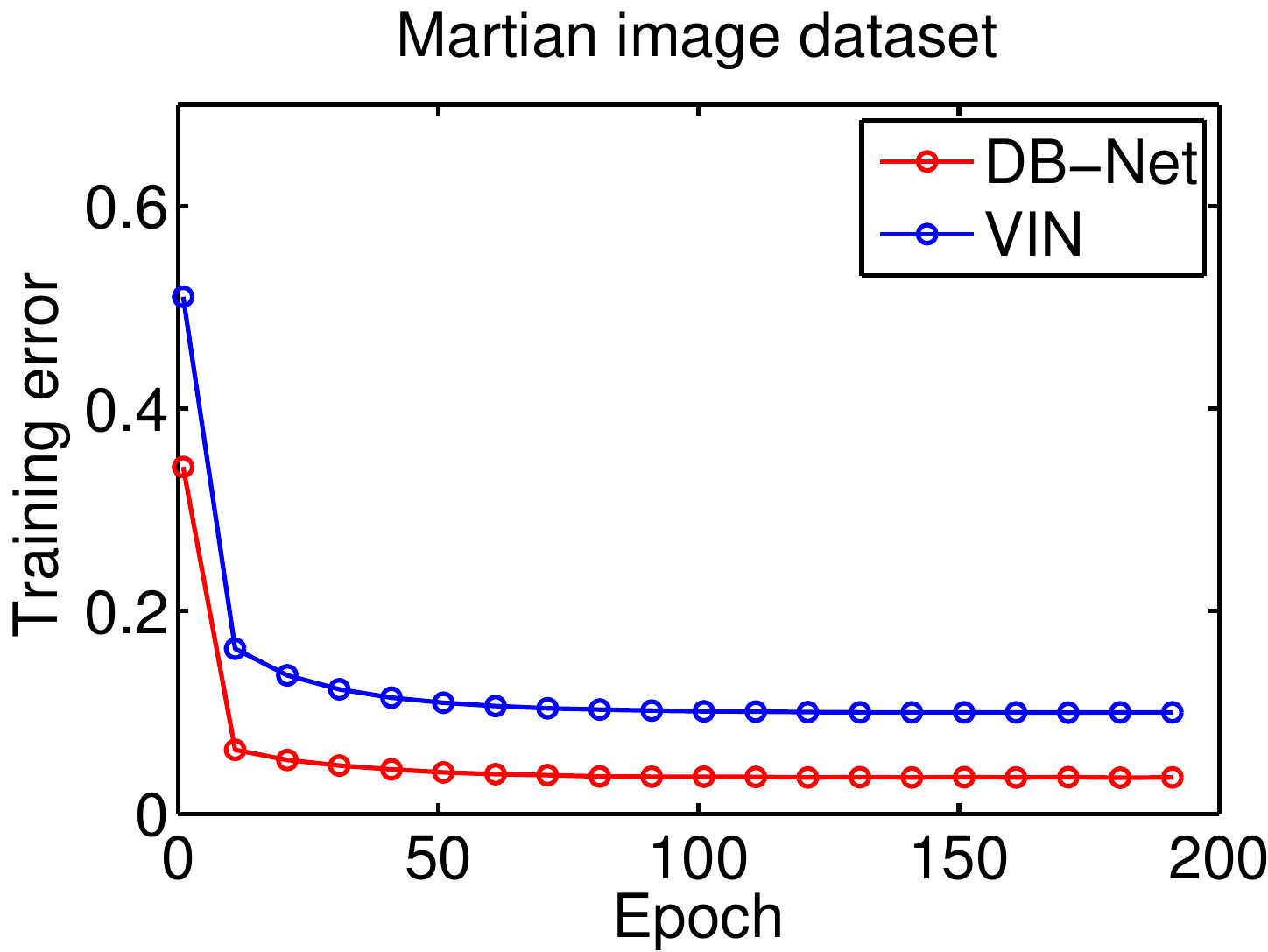}}
\caption{Training results of \emph{DB-Net} and \emph{VIN}}\label{Fig.4.1}
\end{figure}

As illustrated in Fig.~\ref{Fig.4.1}, both training loss and training error of \emph{DB-Net} converge faster that \emph{VIN}. After 200 training epoches, \emph{DB-Net} achieves 96.4\% training accuracy and 95.4\% testing accuracy, outperforming \emph{VIN} significantly in precision (as shown in TABLE~\ref{tab4.1}). Moreover, compared with \emph{VIN}, average time cost of \emph{DB-Net} in one training epoch is reduced by 45.8\%, exceeding \emph{VIN} in efficiency promptly. Finally, \emph{DB-Net} achieves high success rate both in training data and testing data. Remarkably, Martian environment images in testing data are totally unknown to \emph{DB-Net}, since training data differs from testing data. Therefore, even if the environment is unknown before, \emph{DB-Net} can still achieve high-precision visual navigation. By contrast, \emph{VIN} exhibits poor performance on success rate, which is less than 80\% in testing data.

\begin{figure*}
\centering
\subfigure[Successful examples]{
\begin{minipage}[b]{0.24\textwidth}
\centering
\includegraphics[width=1\linewidth]{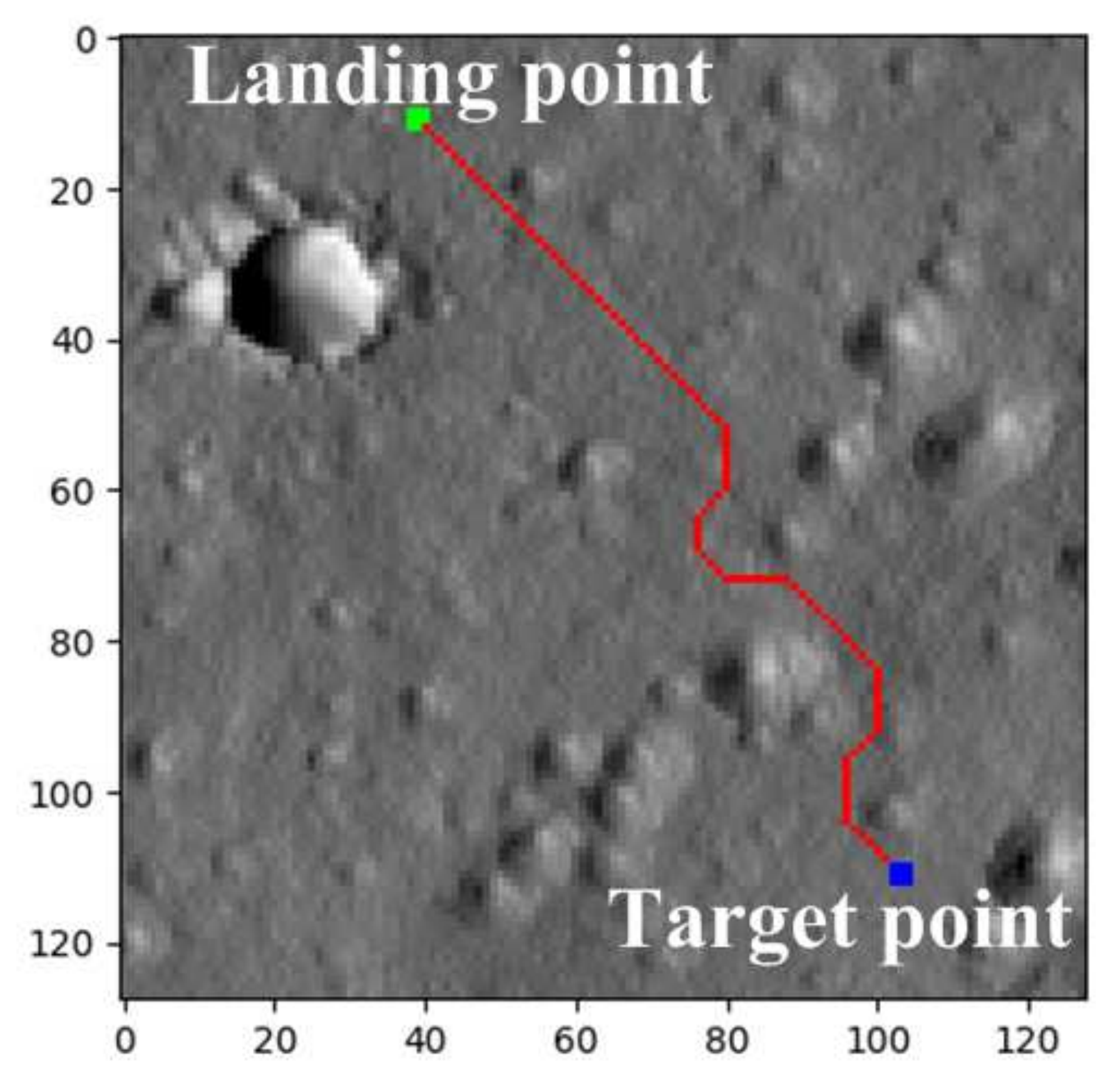}\\
\includegraphics[width=1\linewidth]{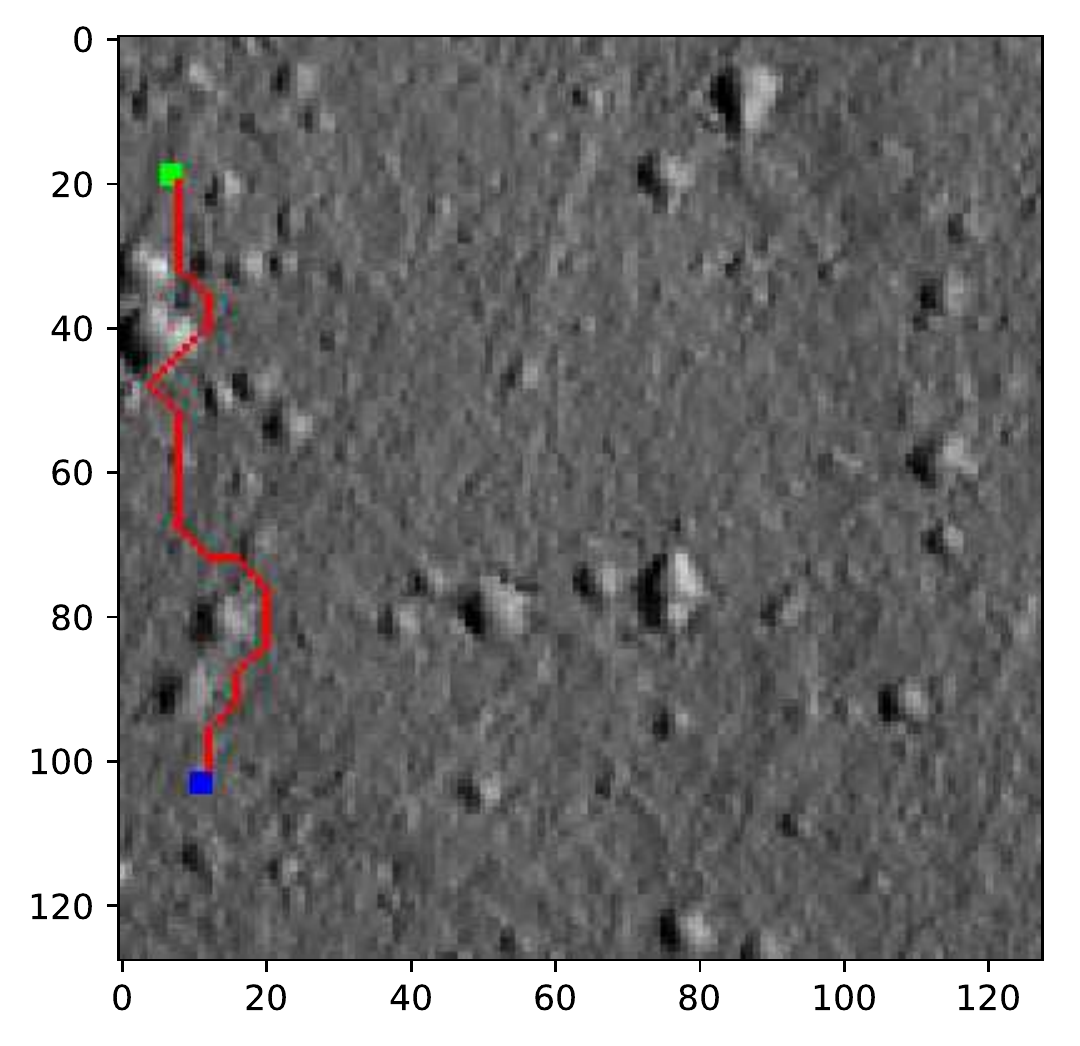}\\
\includegraphics[width=1\linewidth]{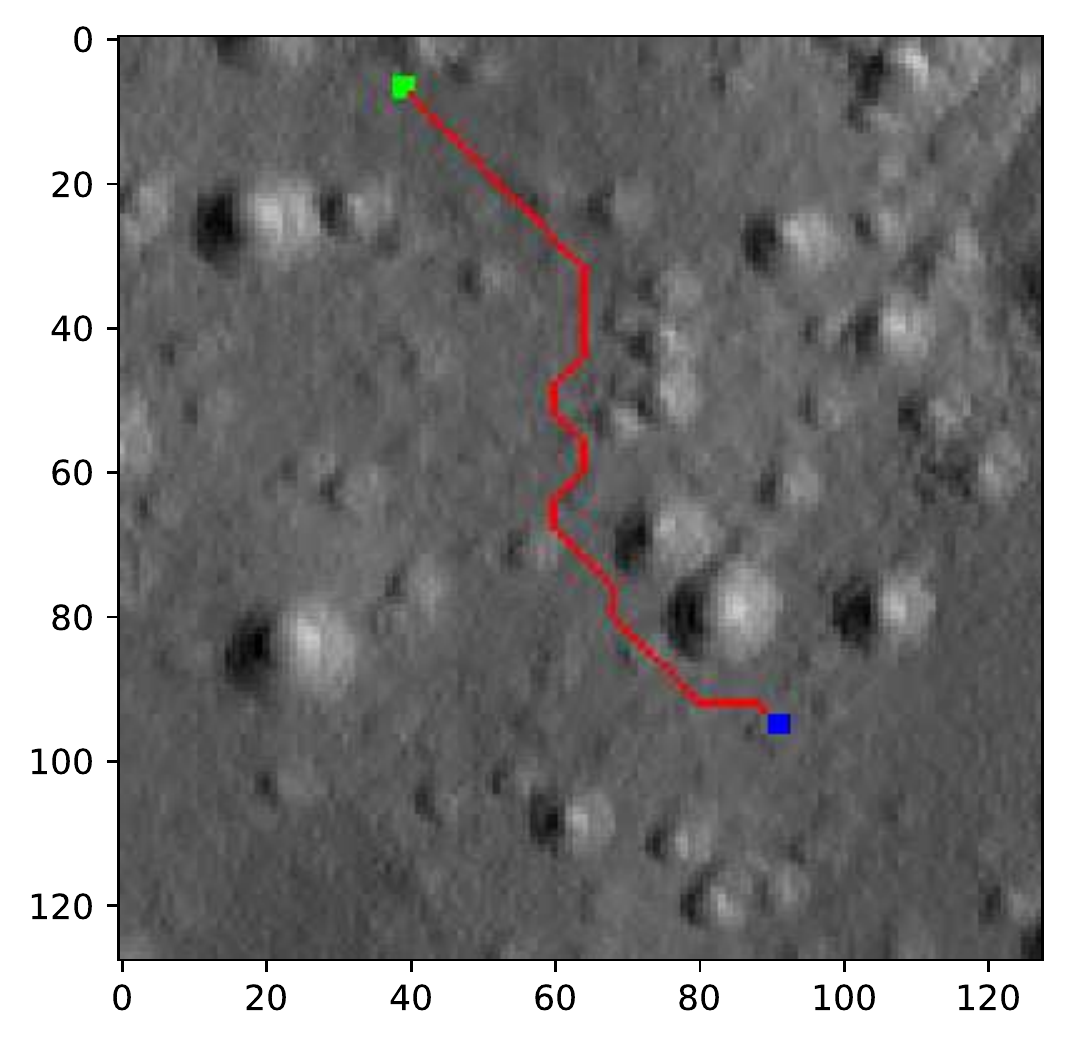}
\end{minipage}
\begin{minipage}[b]{0.24\textwidth}
\centering
\includegraphics[width=1\linewidth]{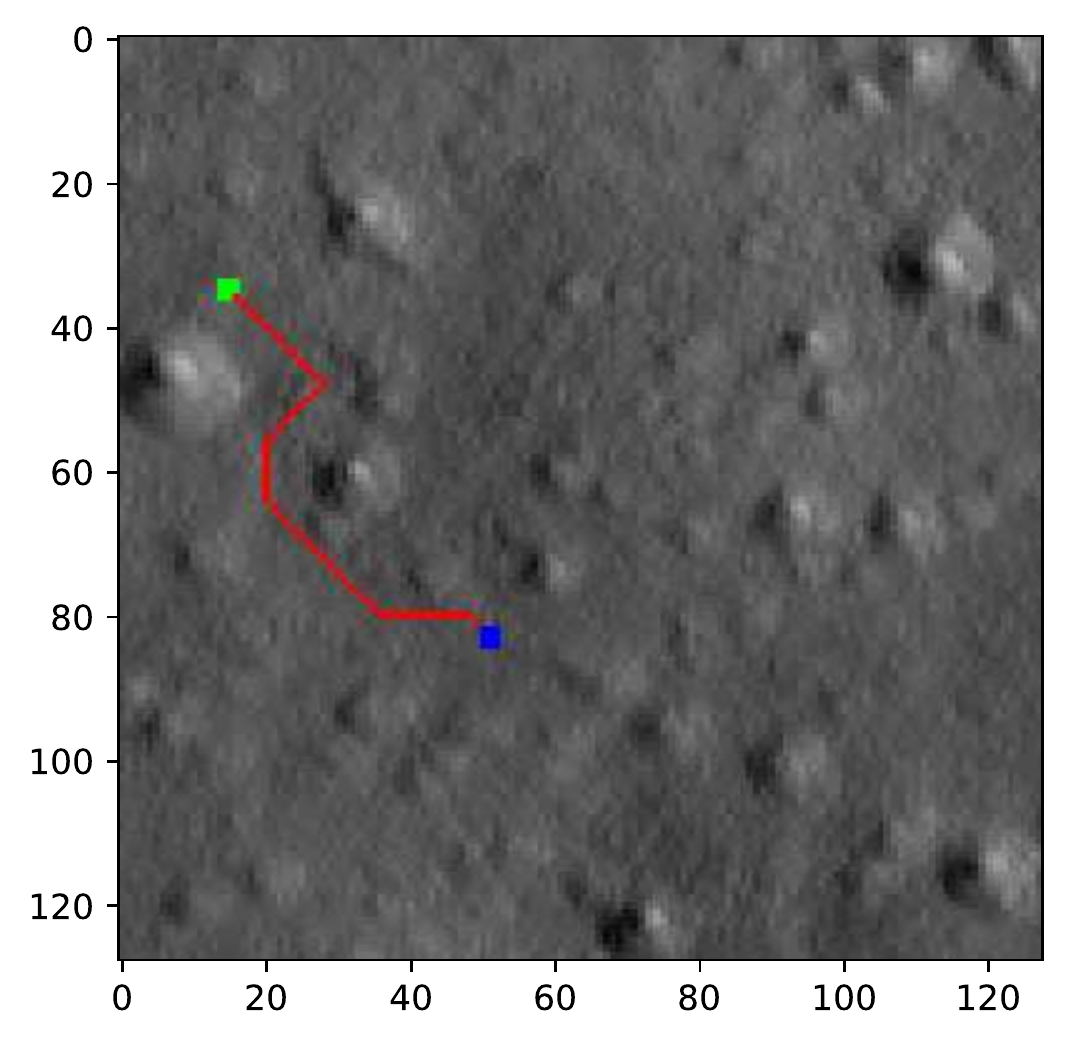}\\
\includegraphics[width=1\linewidth]{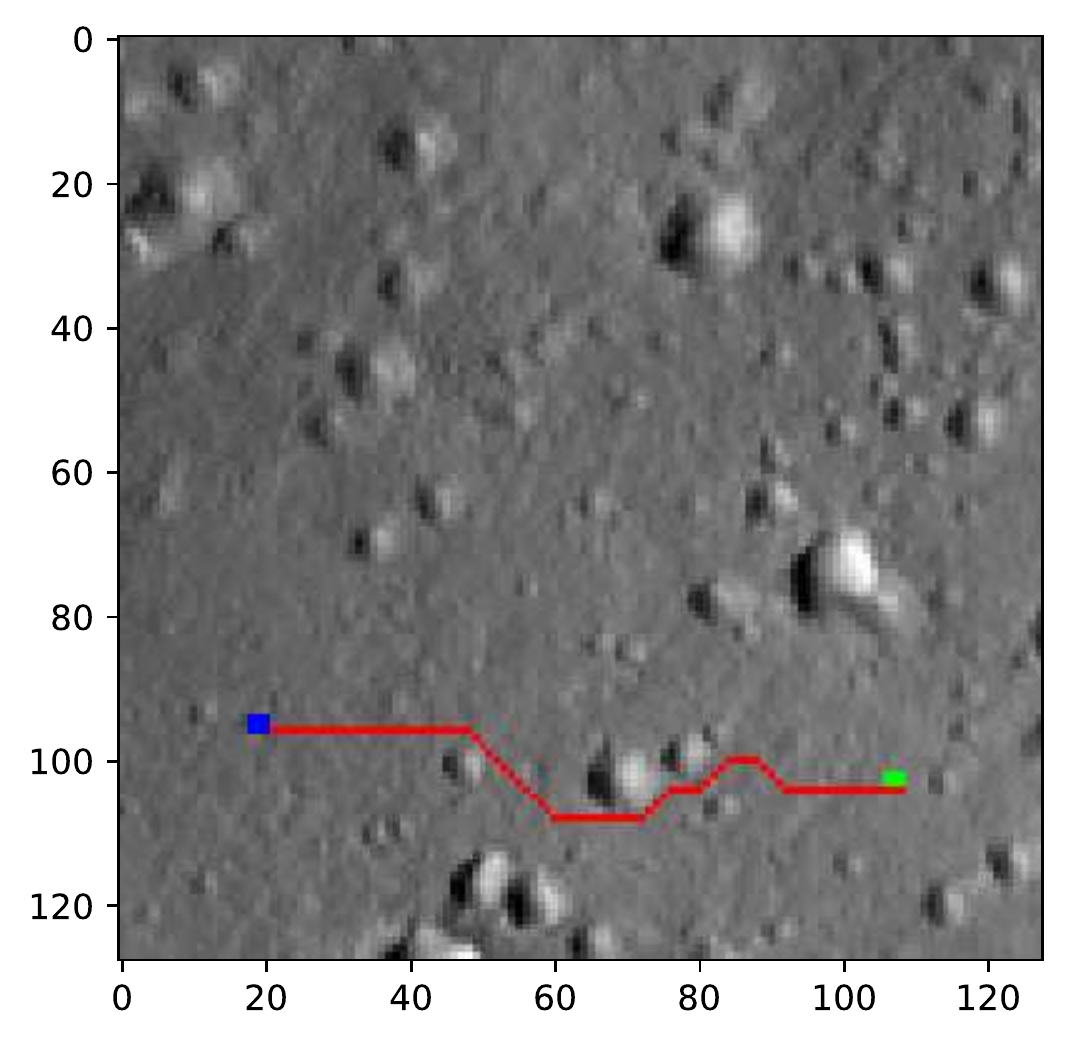}\\
\includegraphics[width=1\linewidth]{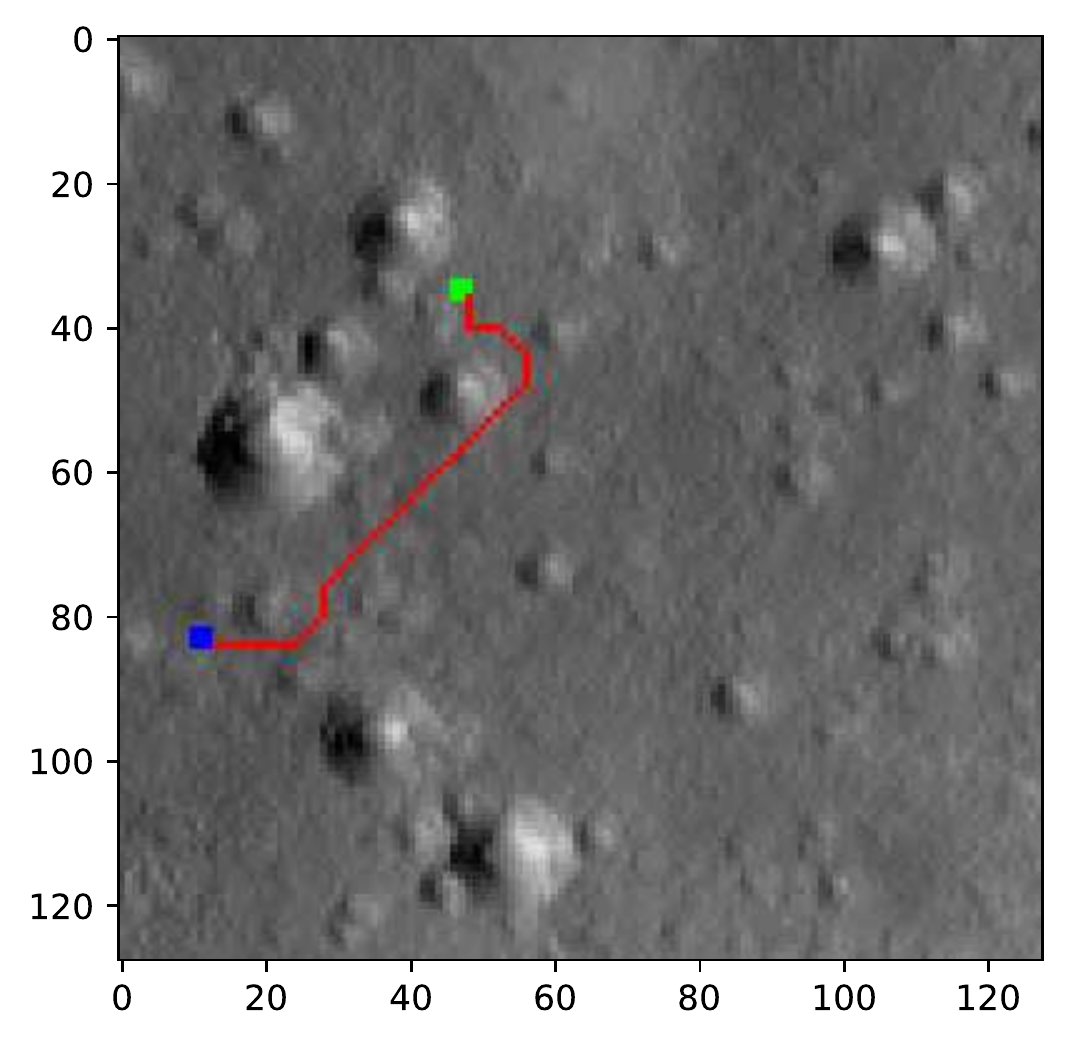}
\end{minipage}
\begin{minipage}[b]{0.24\textwidth}
\centering
\includegraphics[width=1\linewidth]{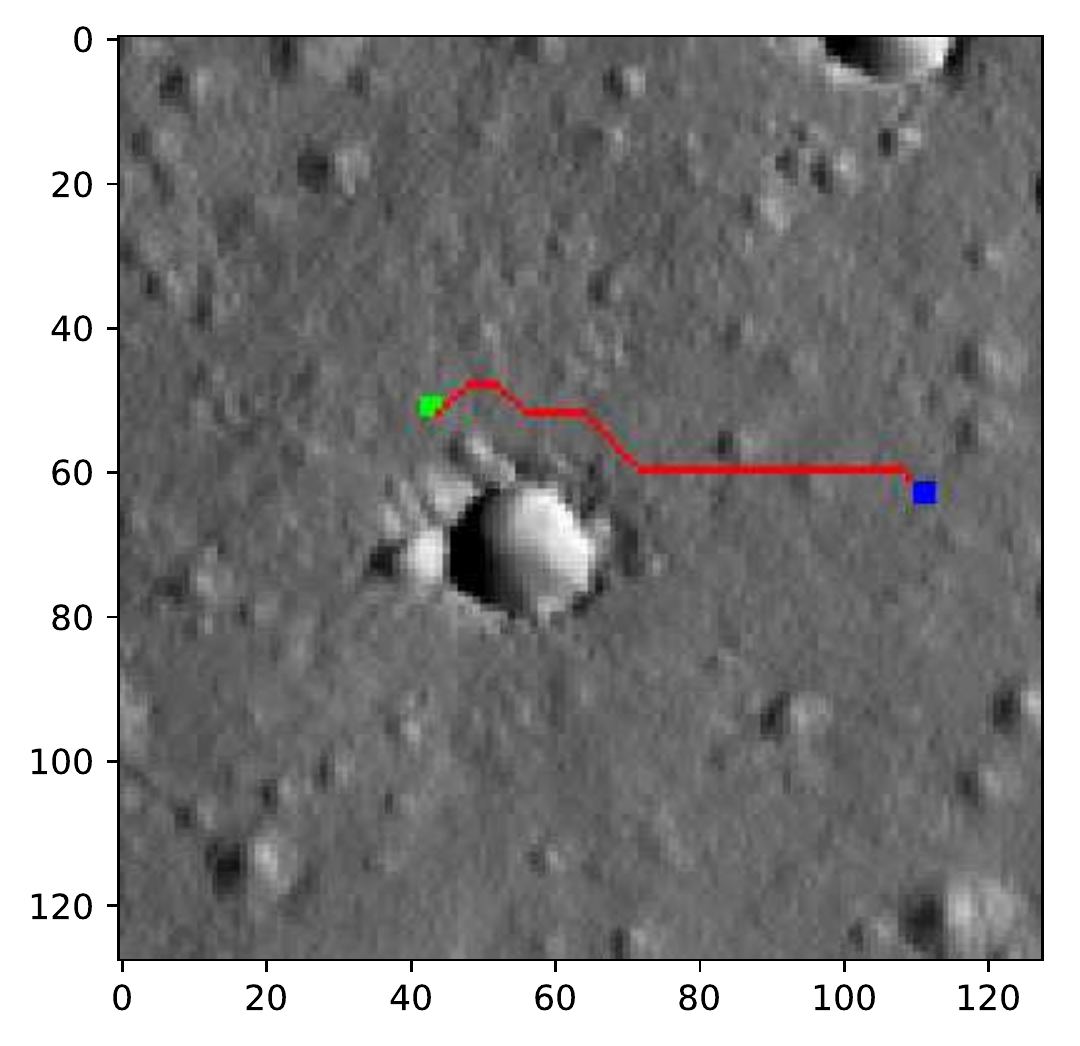}\\
\includegraphics[width=1\linewidth]{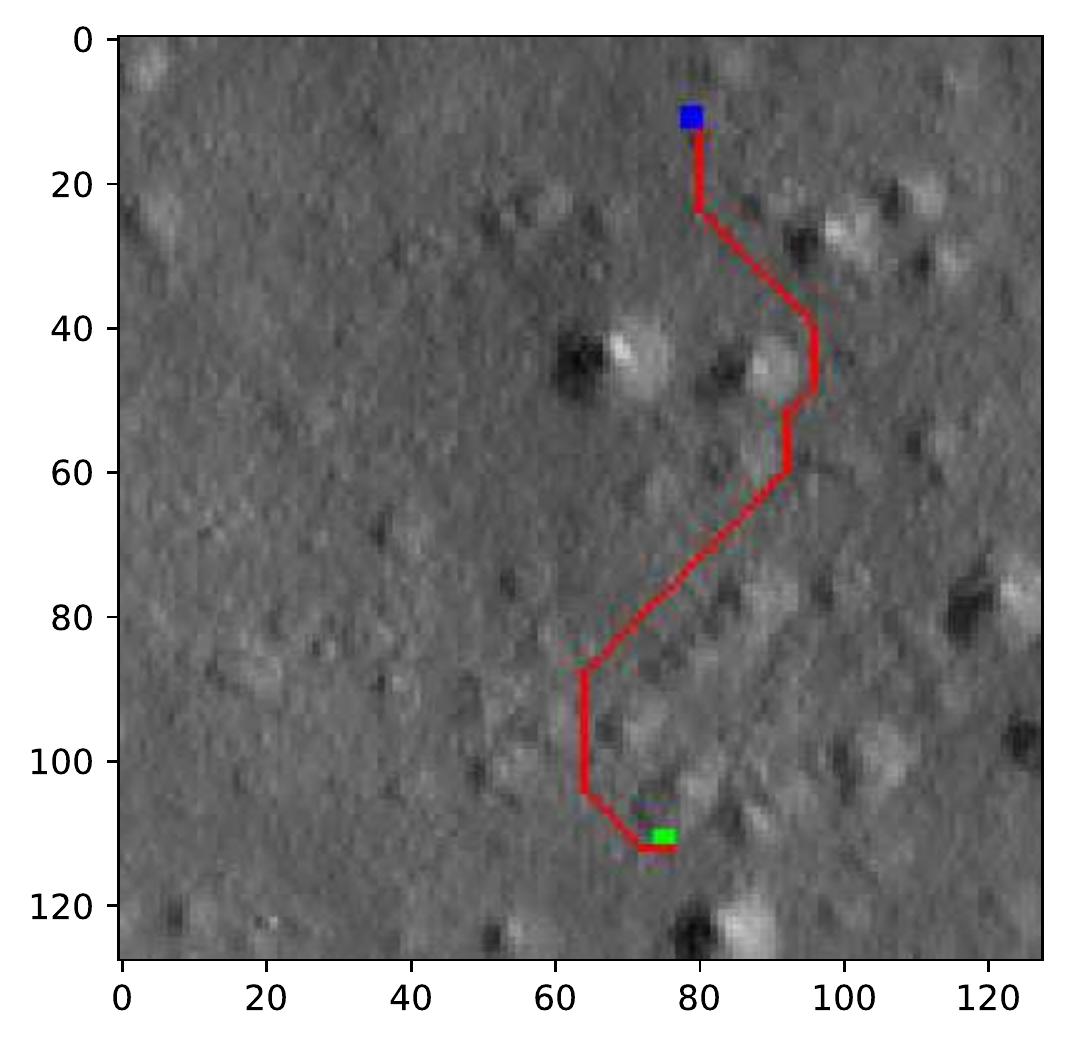}\\
\includegraphics[width=1\linewidth]{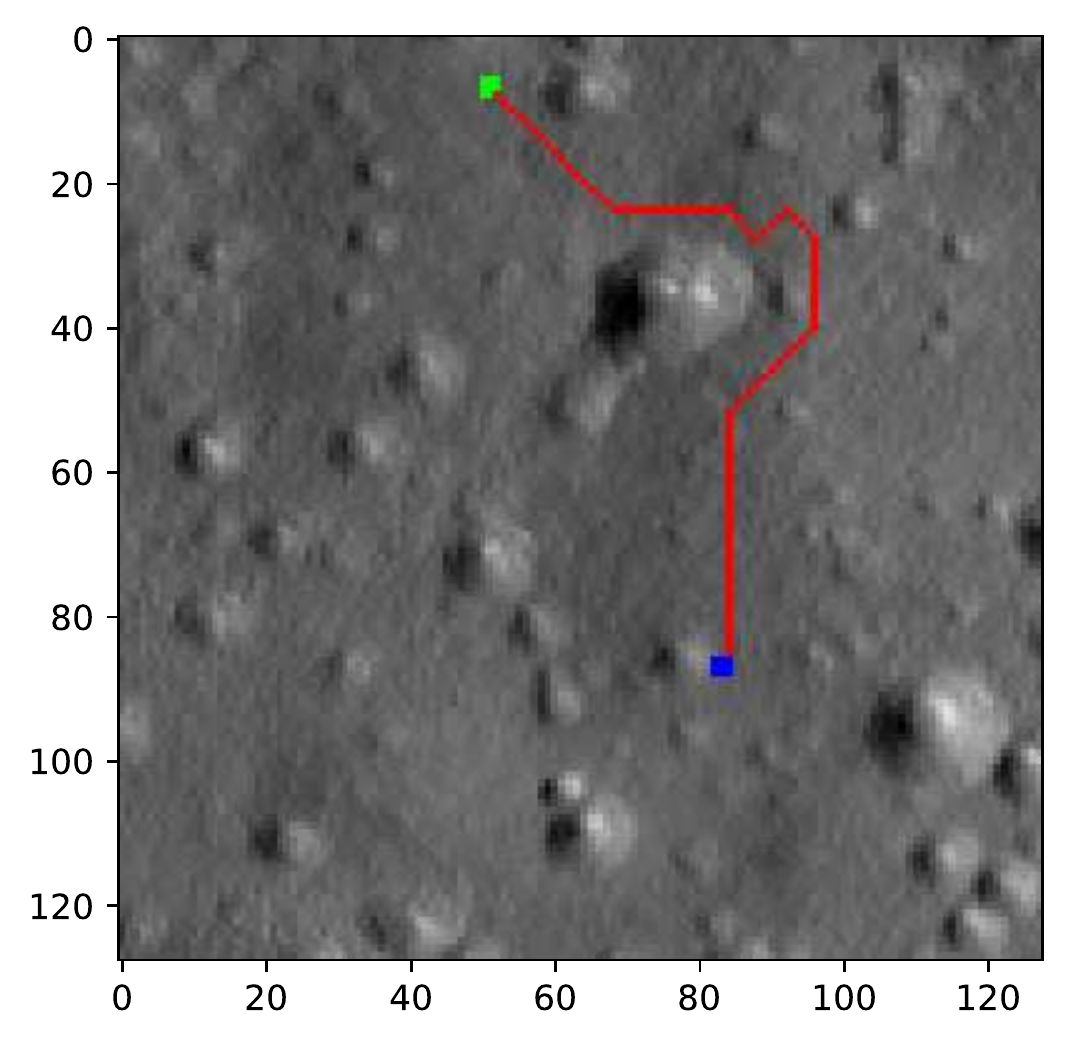}
\end{minipage}
\begin{minipage}[b]{0.24\textwidth}
\centering
\includegraphics[width=1\linewidth]{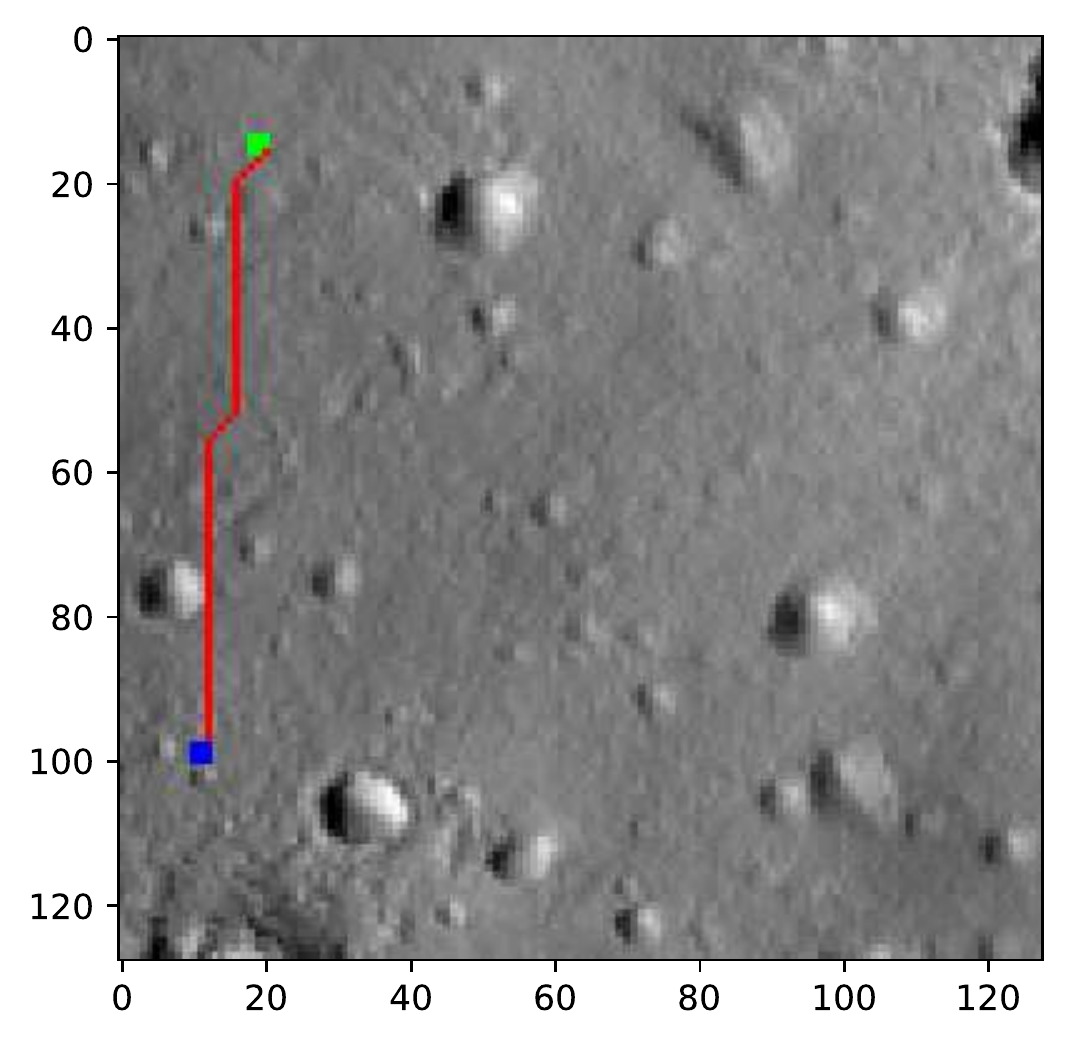}\\
\includegraphics[width=1\linewidth]{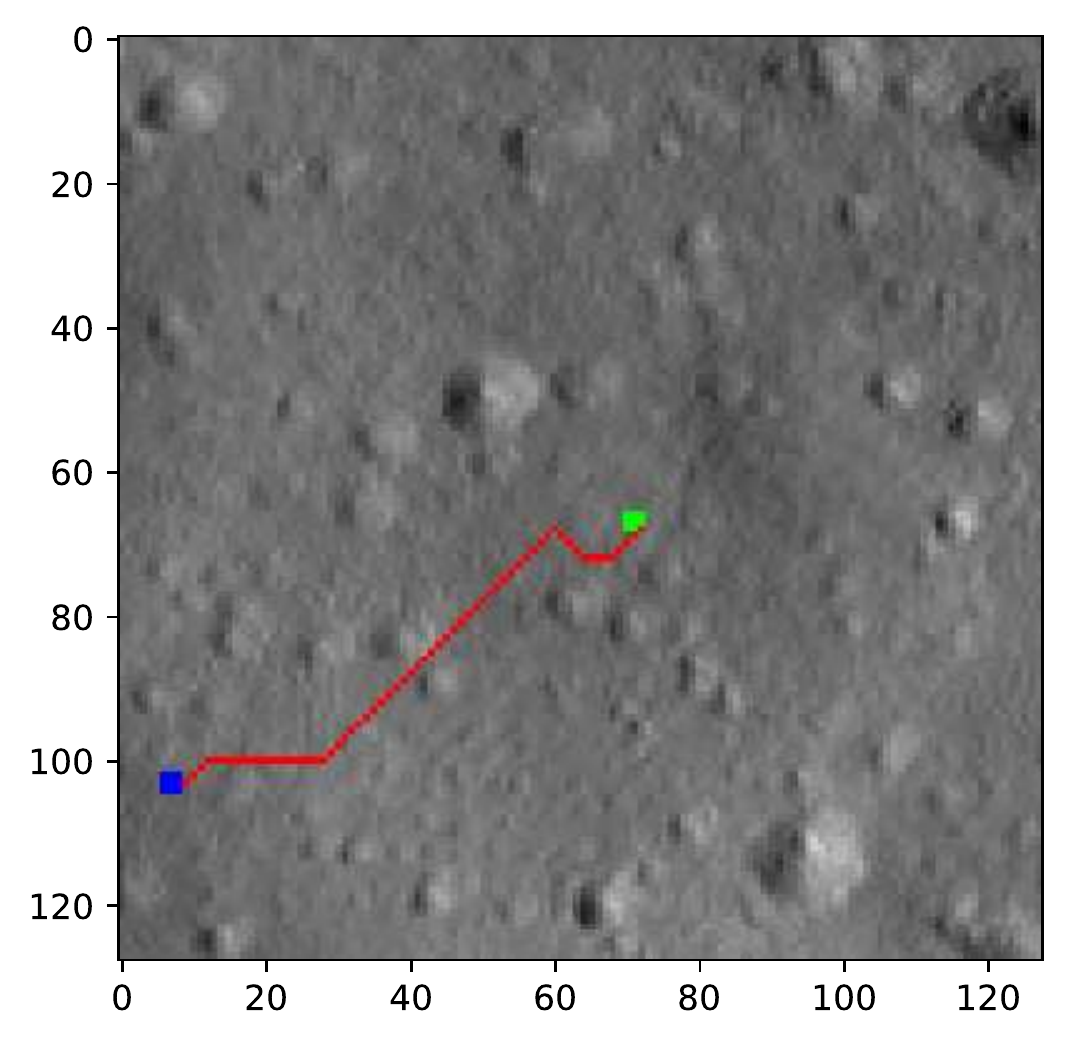}\\
\includegraphics[width=1\linewidth]{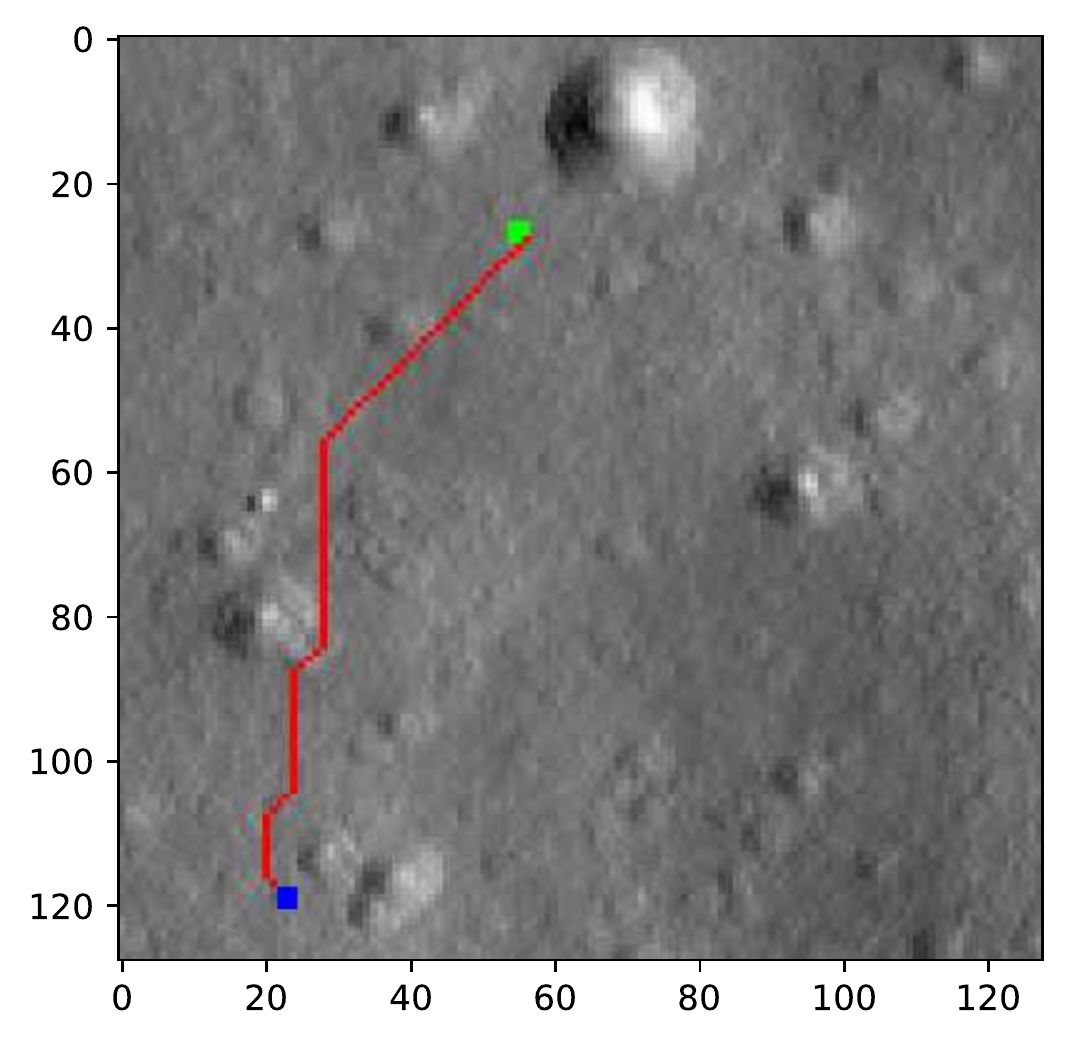}
\end{minipage}}
\subfigure[Failed examples]{
\begin{minipage}[b]{0.24\textwidth}
\centering
\includegraphics[width=1\linewidth]{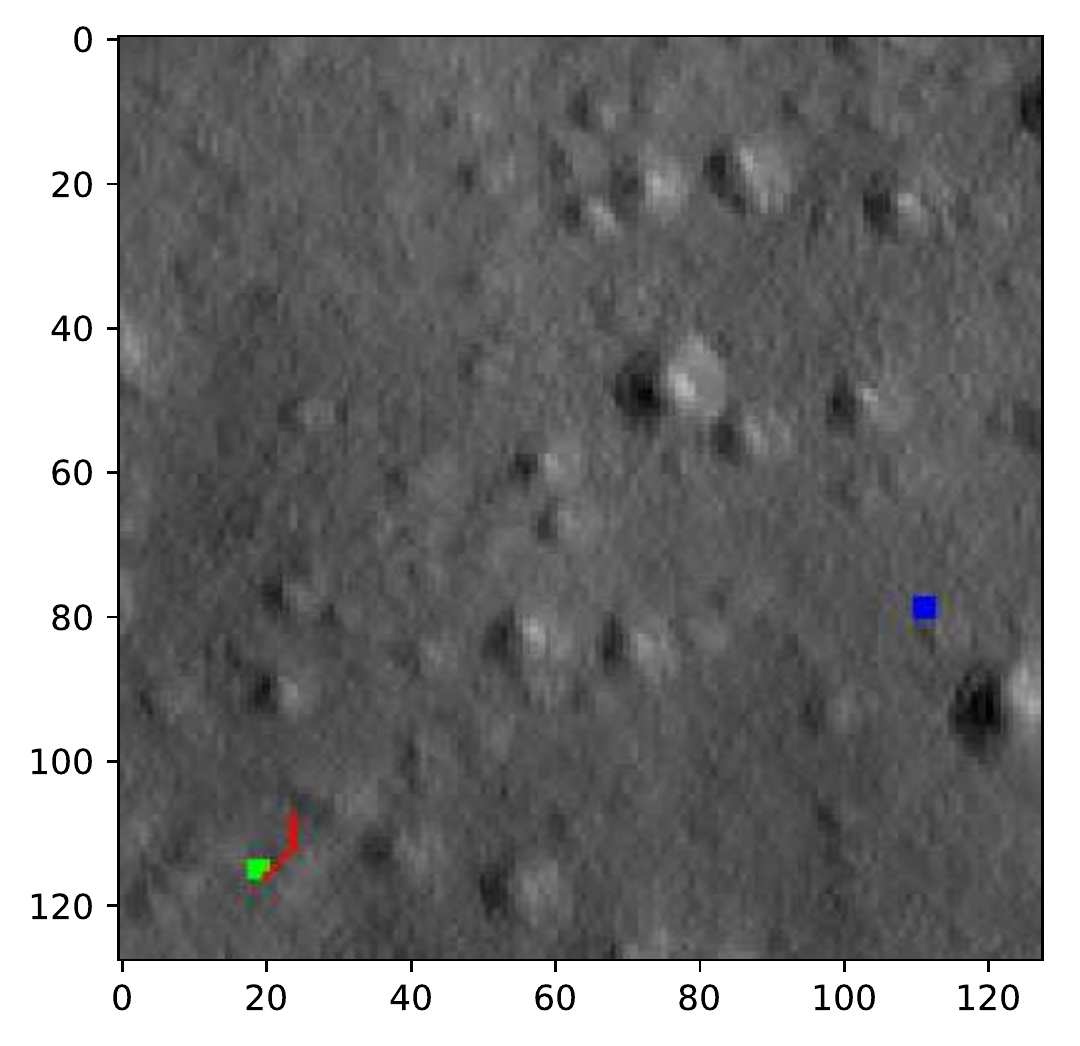}
\end{minipage}
\begin{minipage}[b]{0.24\textwidth}
\centering
\includegraphics[width=1\linewidth]{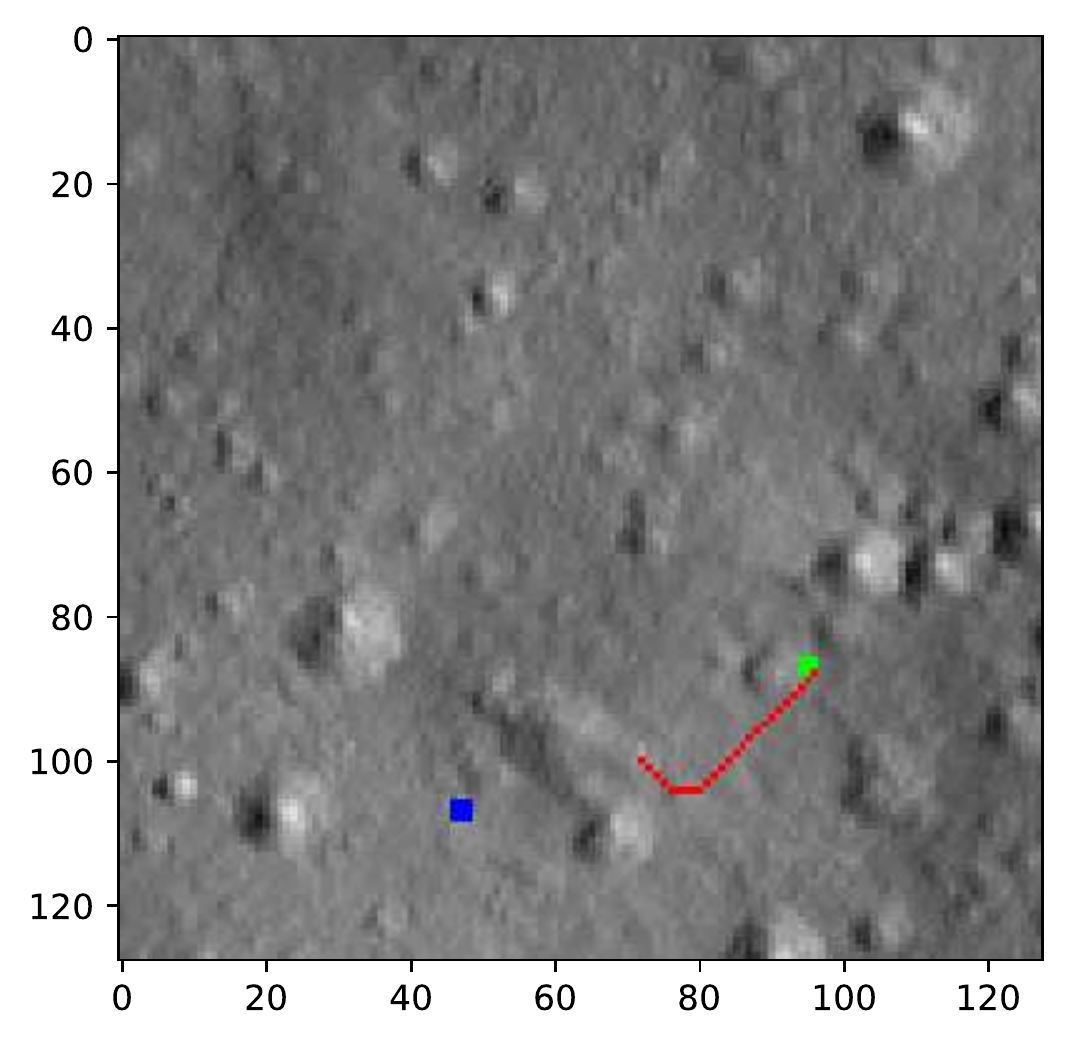}
\end{minipage}
\begin{minipage}[b]{0.24\textwidth}
\centering
\includegraphics[width=1\linewidth]{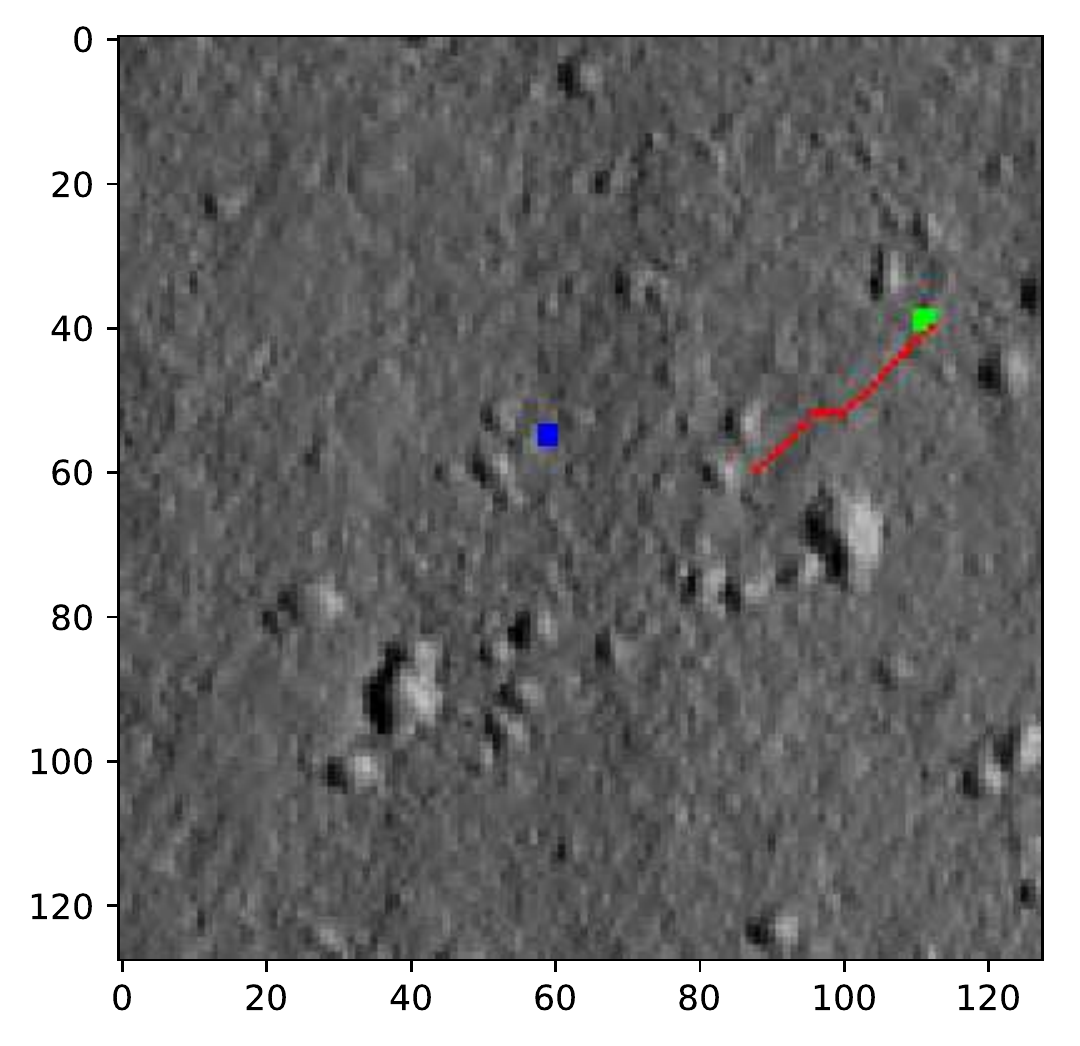}
\end{minipage}
\begin{minipage}[b]{0.24\textwidth}
\centering
\includegraphics[width=1\linewidth]{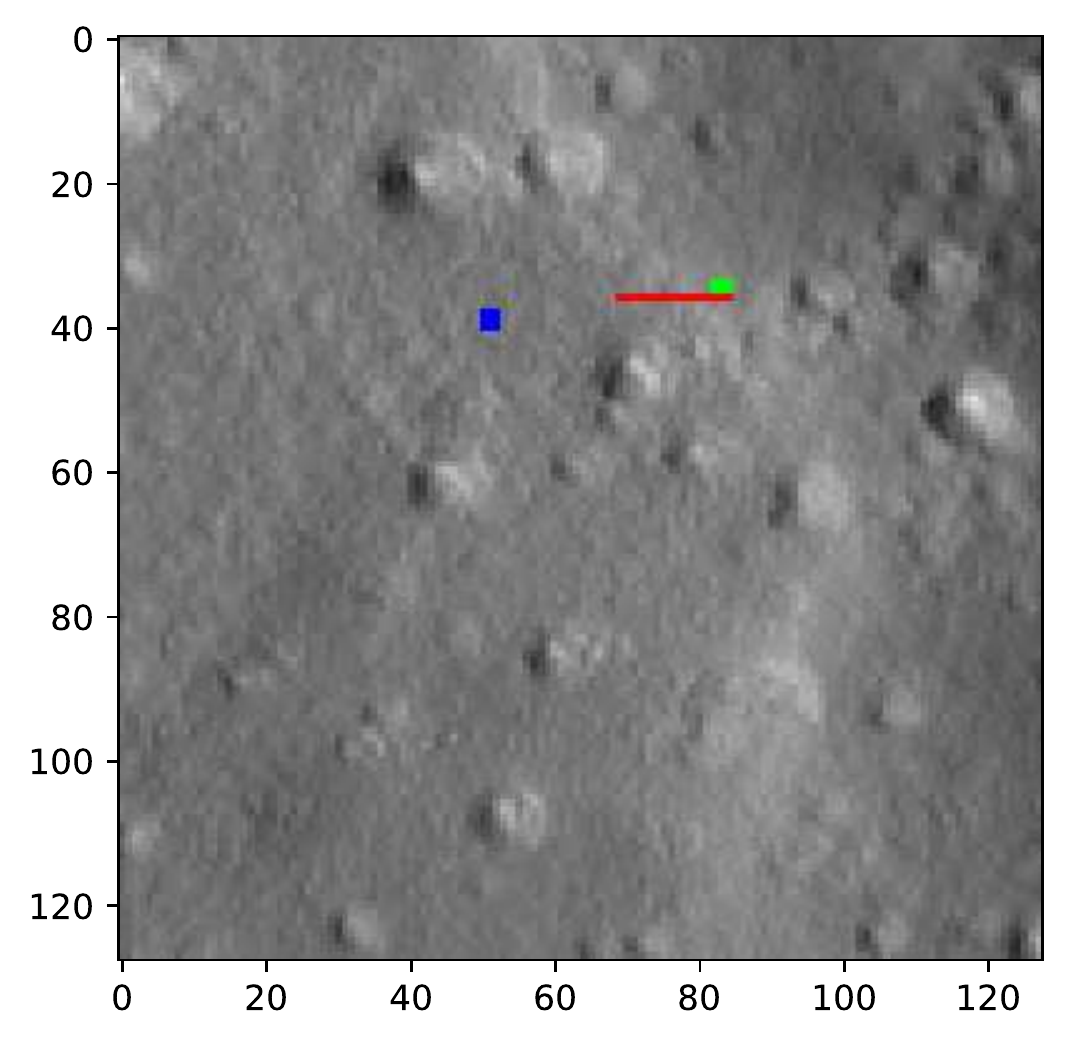}
\end{minipage}}
\caption{Experiments on $128\times128$ Martian Images. (Green points are landing points. Blue points are the target points. Navigation trajectories are red.)}\label{Fig.4.2}
\end{figure*}

Examples of successful navigation process are demonstrated in Fig.~\ref{Fig.4.2}. It can be seen that the rover avoid craters with varying size precisely under the guidance of \emph{DB-Net}. Furthermore, the trajectories are nearly optimal. It is worth noting that prior knowledge of craters are unknown and \emph{DB-Net} has to understand deep representations of original Martian images intuitively. Therefore, the performance of \emph{DB-Net} is marvellous.
\begin{table}[H]
\centering
\caption{Results on 128x128 Martian image}\label{tab4.1}
\begin{tabular}{|c|c|c|c|c|c|c|}
\hline
Architectures                     & \emph{DB-Net}     & \emph{VIN}  \\\hline
Training accuracy                 & \textbf{96.4\%}   & 90.0\%      \\\hline
Testing accuracy                  & \textbf{95.6\%}   & 89.8\%      \\\hline
Training success rate             & \textbf{96.0\%}   & 81.1\%      \\\hline
Testing success rate              & \textbf{93.3\%}   & 79.4\%      \\\hline
Average time cost (each epoch)    & \textbf{52.8s}    & 97.5s       \\\hline
\end{tabular}
\end{table}

\subsection{Model Ablation Analysis of \emph{DB-Net}}
\begin{figure}[H]
\centering
\subfigure[Training loss]{
\includegraphics[width=0.47\linewidth]{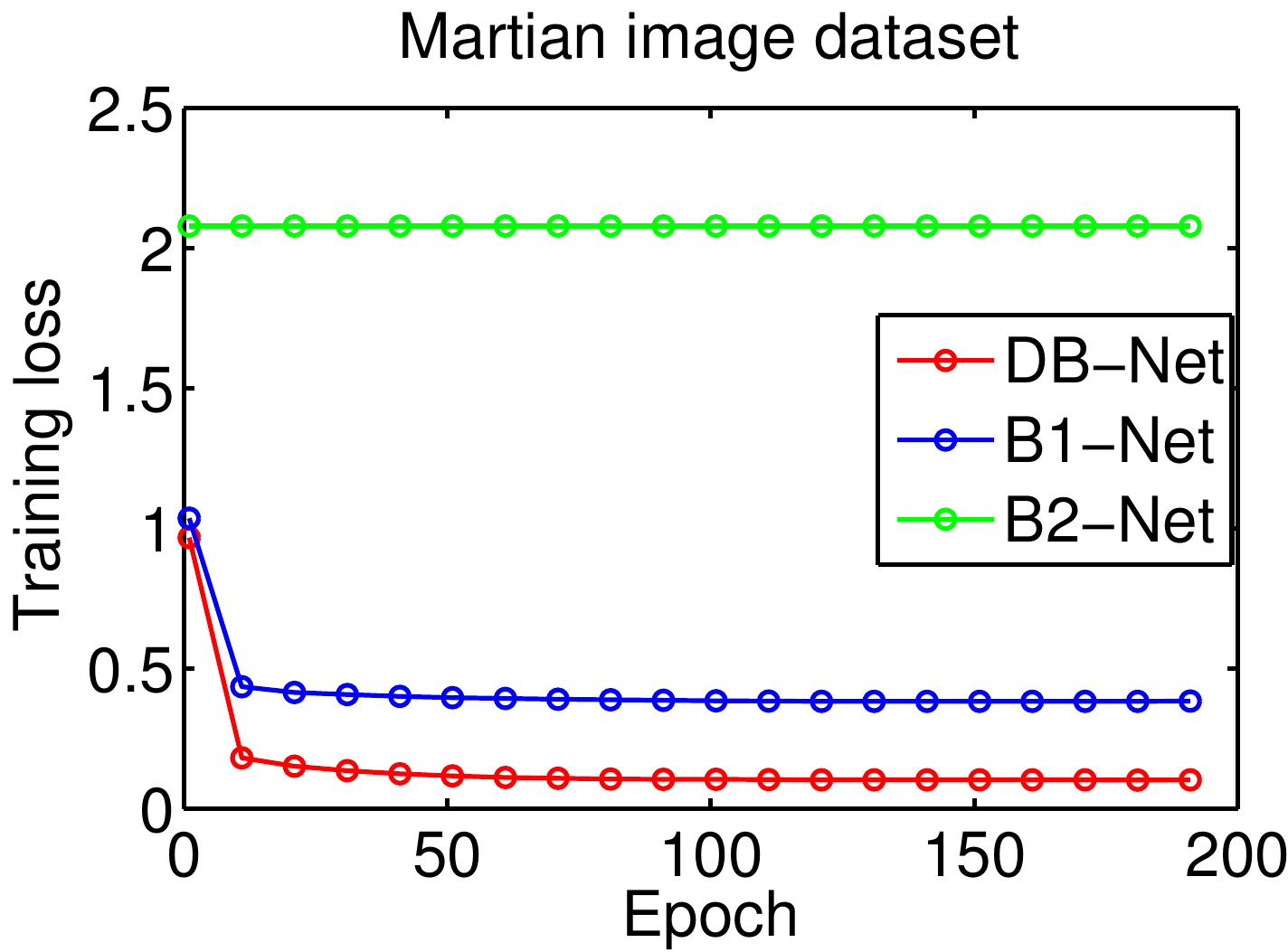}}
\subfigure[Training error]{
\includegraphics[width=0.47\linewidth]{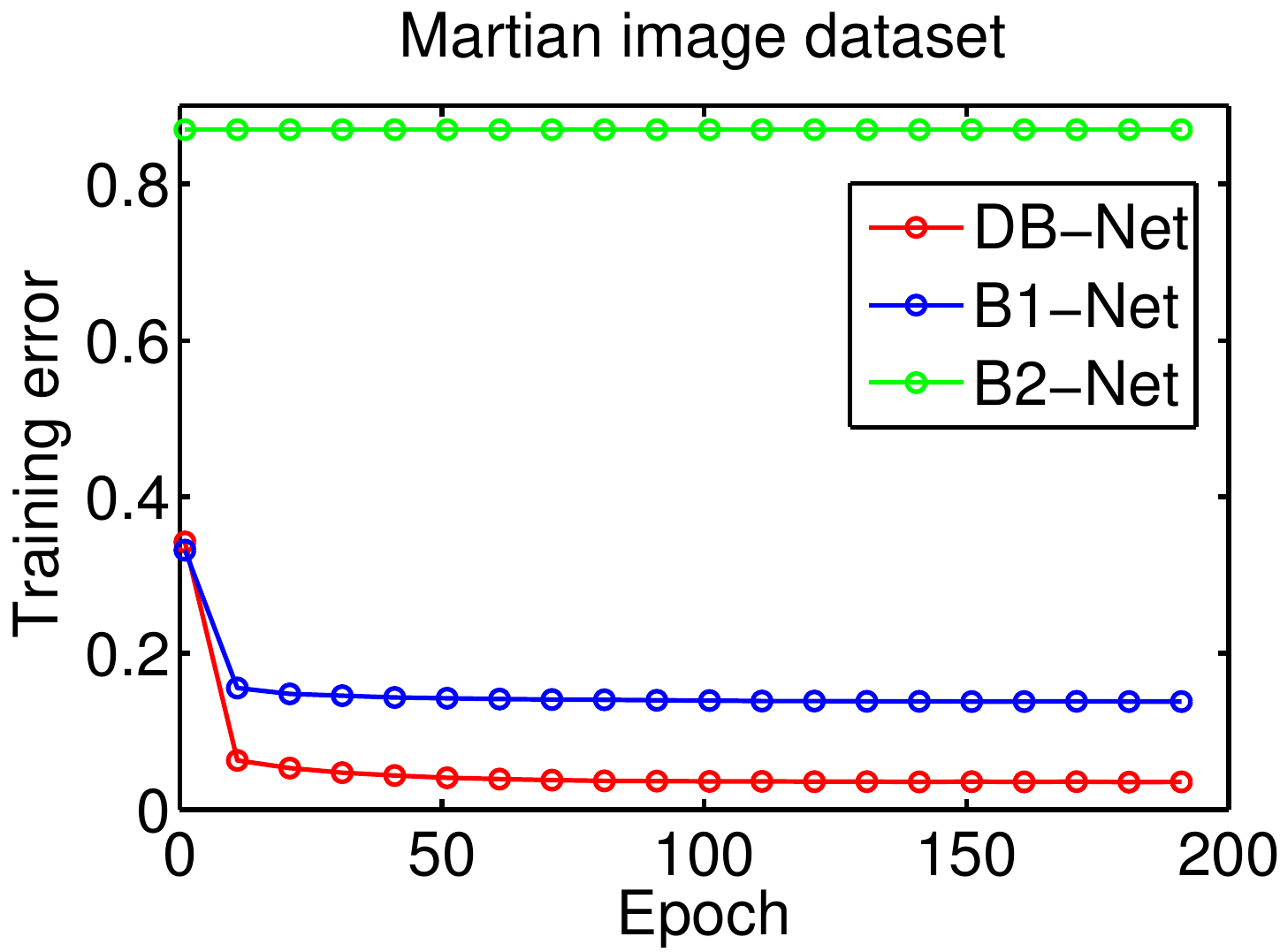}}
\caption{Training results of \emph{DB-Net}, \emph{B1-Net} and \emph{B2-Net}}\label{Fig.4.3}
\end{figure}

\begin{figure*}
\centering
\subfigure[Original Martian images]{
\begin{minipage}[c]{0.24\textwidth}
\centering
\includegraphics[width=1\linewidth]{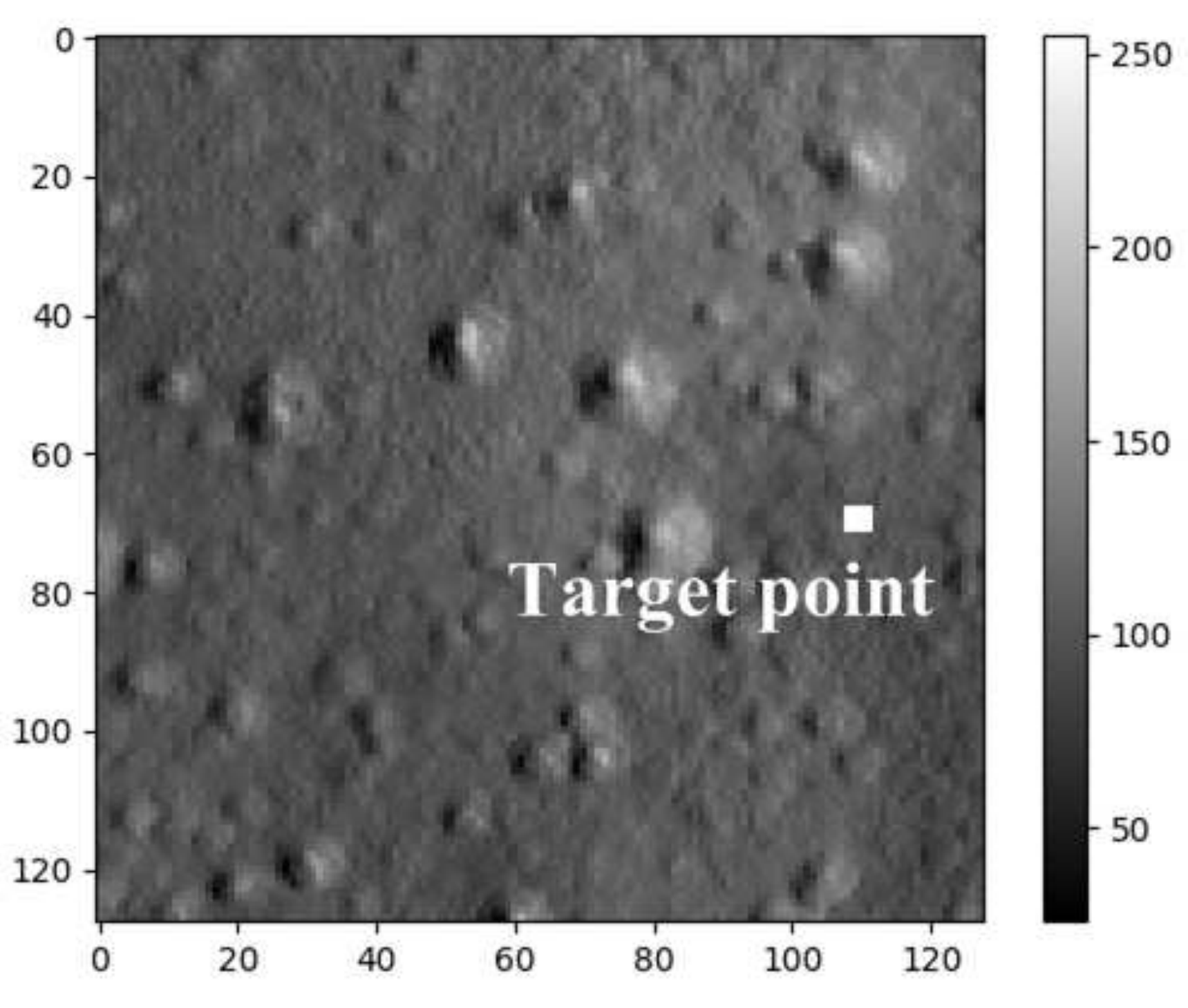}
\end{minipage}
\begin{minipage}[c]{0.24\textwidth}
\centering
\includegraphics[width=1\linewidth]{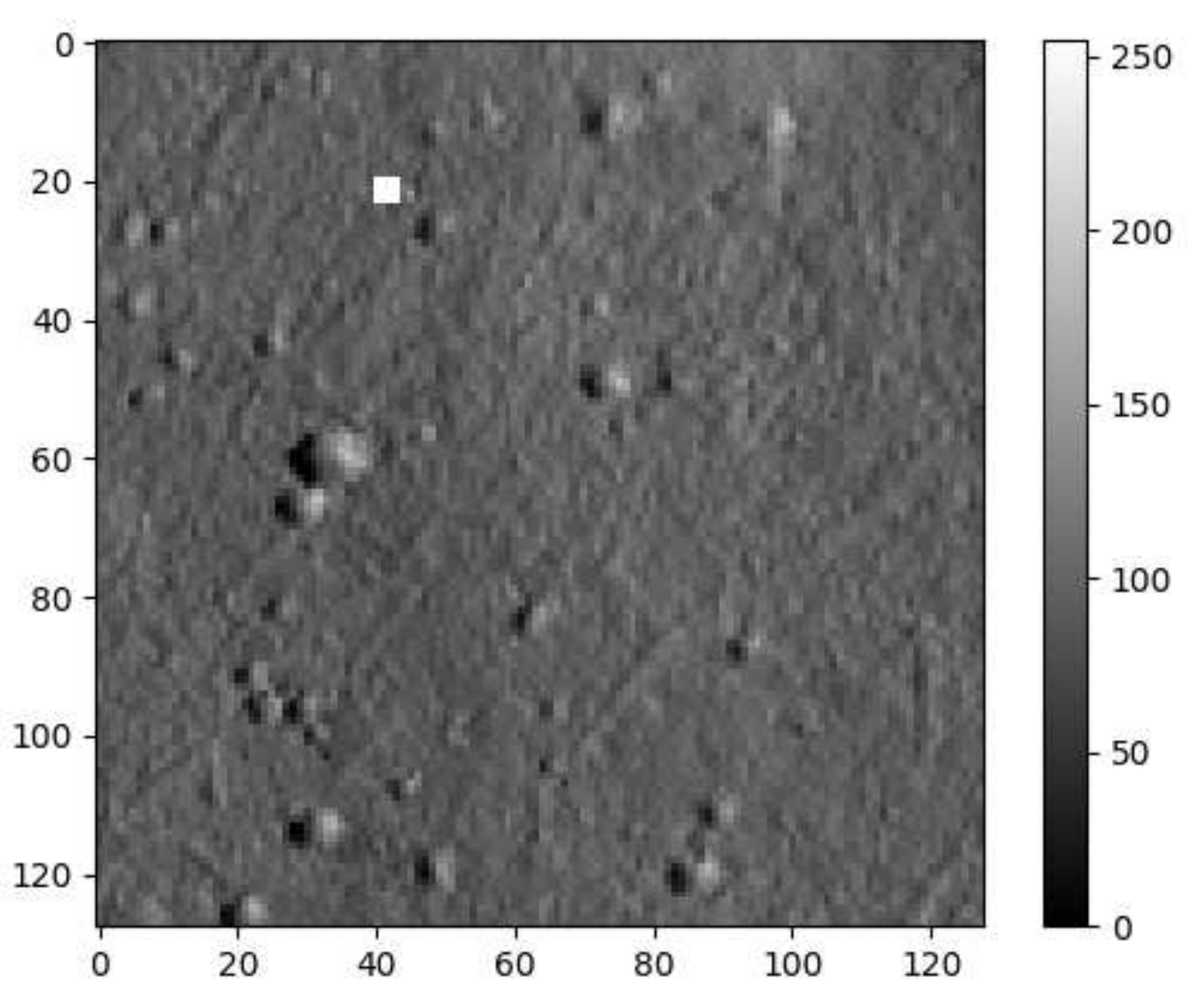}
\end{minipage}
\begin{minipage}[c]{0.24\textwidth}
\centering
\includegraphics[width=1\linewidth]{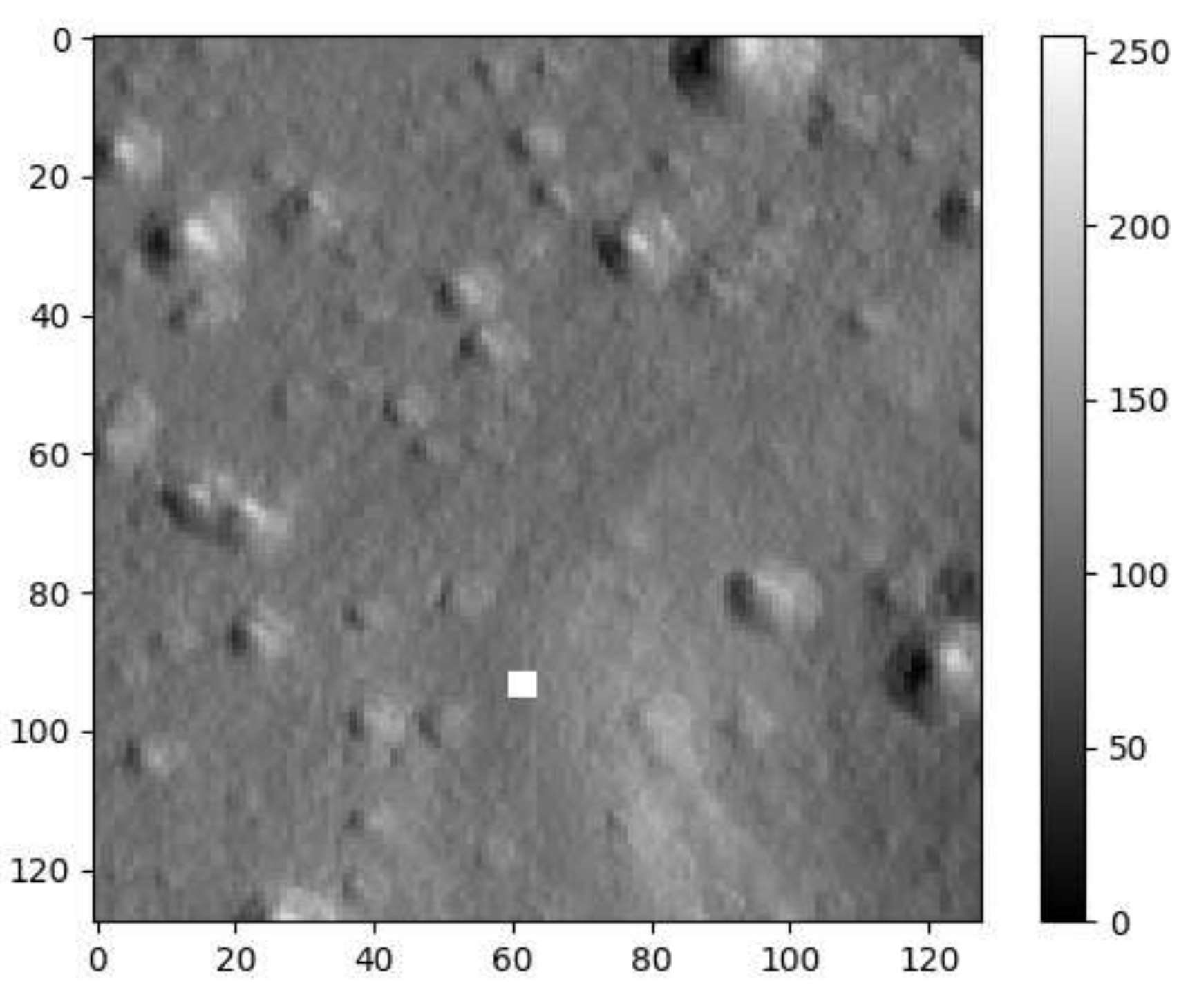}
\end{minipage}
\begin{minipage}[c]{0.24\textwidth}
\centering
\includegraphics[width=1\linewidth]{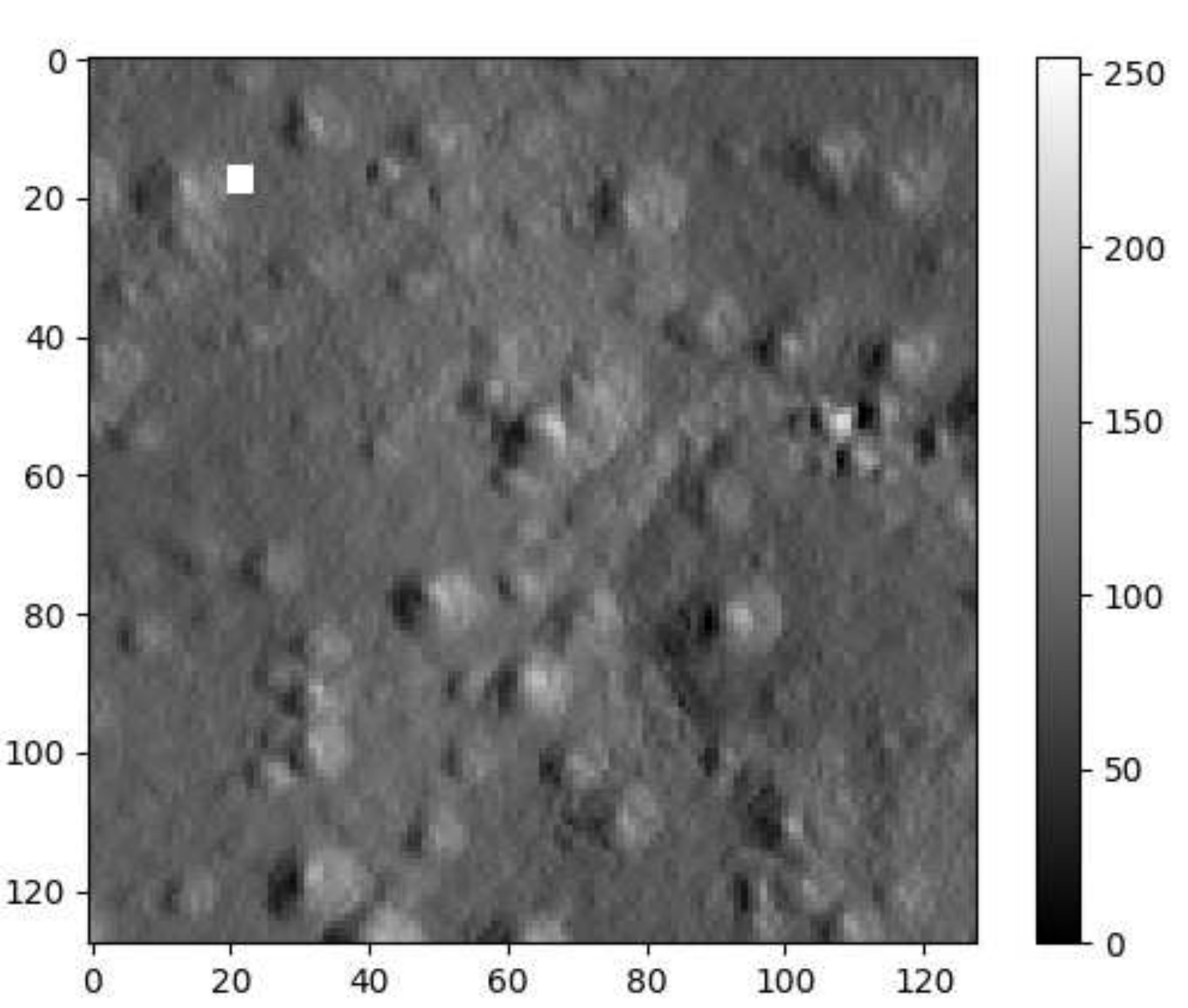}
\end{minipage}}
\subfigure[Value functions estimated by by \emph{DB-Net}]{
\begin{minipage}[c]{0.24\textwidth}
\centering
\includegraphics[width=1\linewidth]{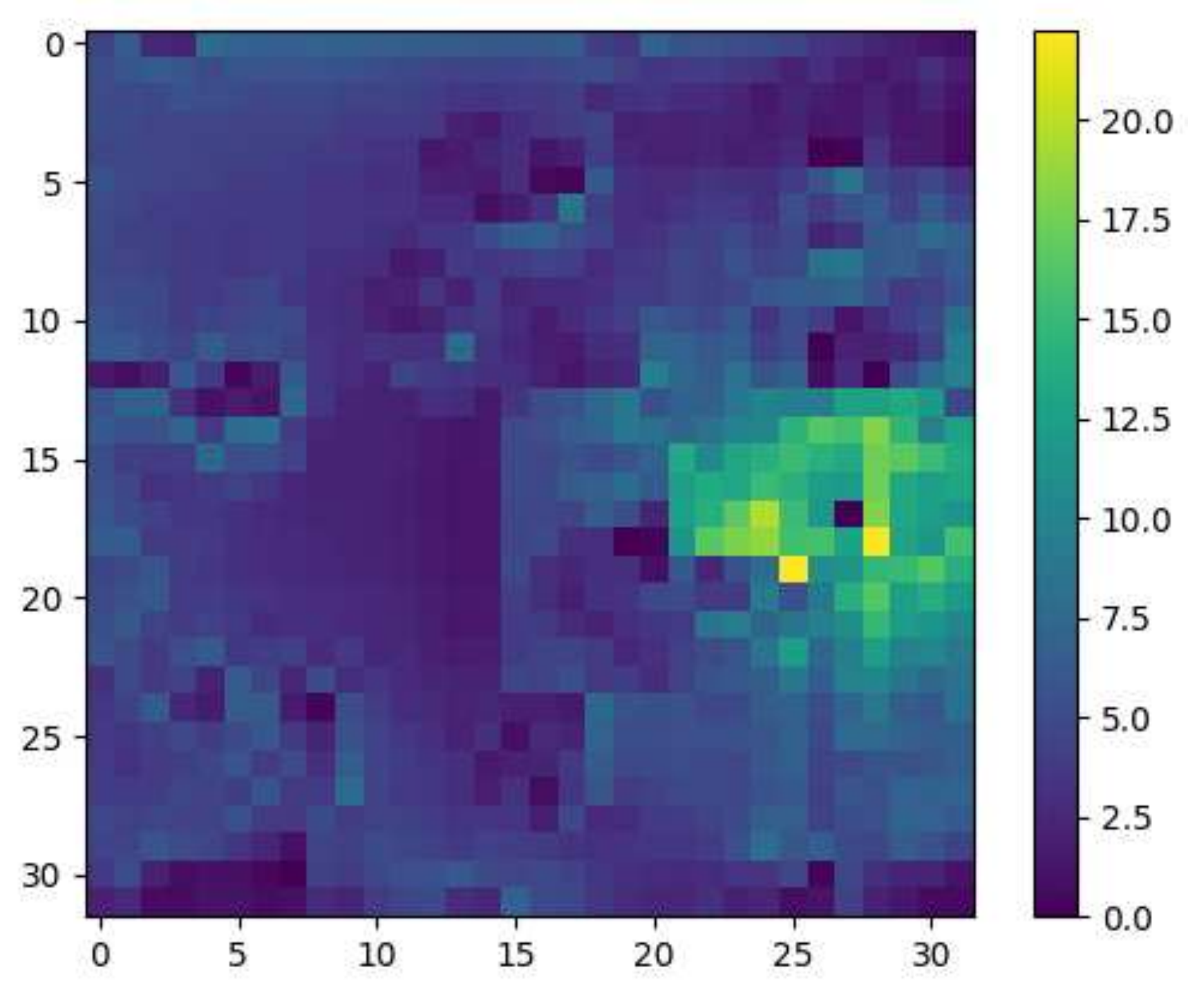}
\end{minipage}
\begin{minipage}[c]{0.24\textwidth}
\centering
\includegraphics[width=1\linewidth]{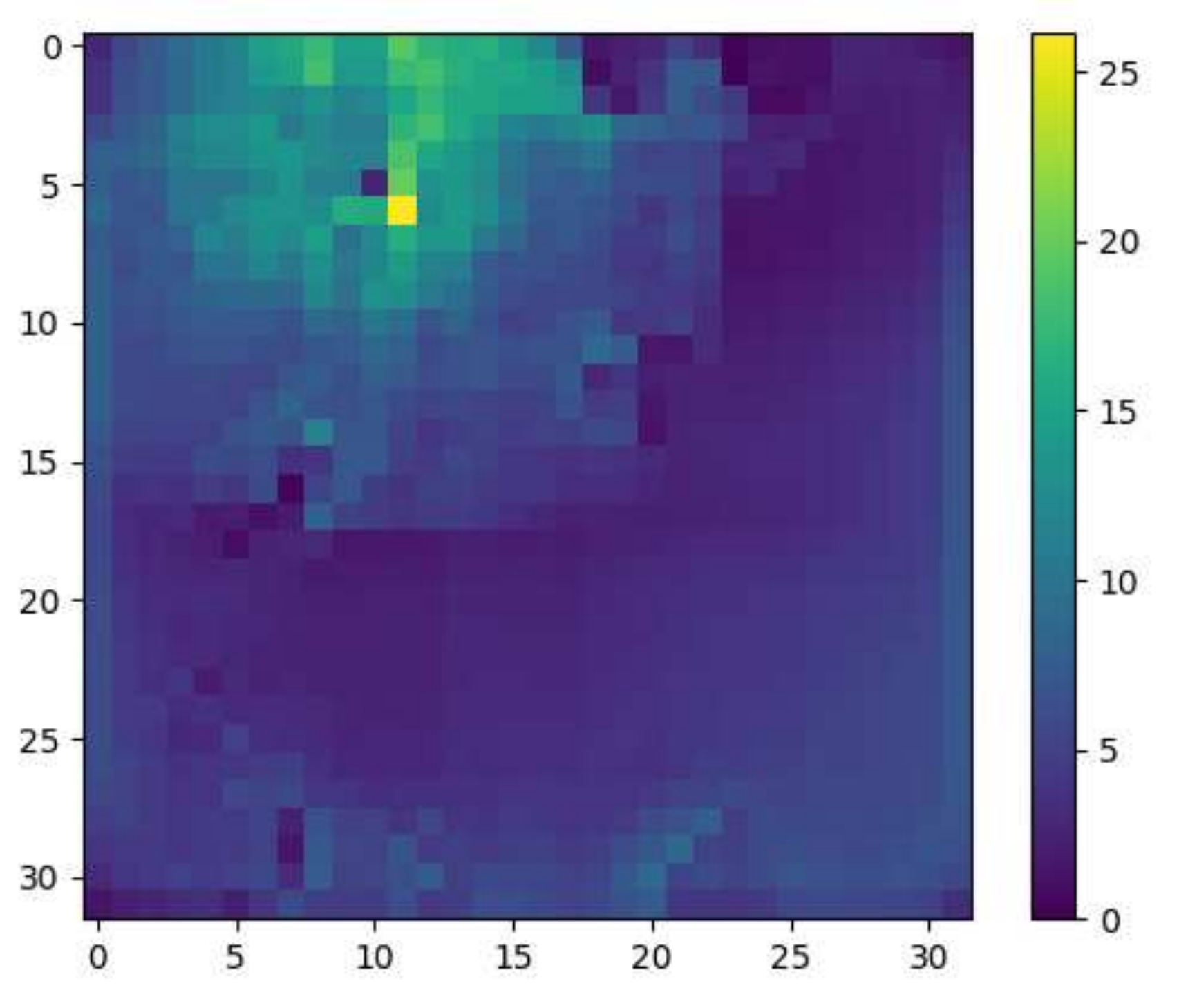}
\end{minipage}
\begin{minipage}[c]{0.24\textwidth}
\centering
\includegraphics[width=1\linewidth]{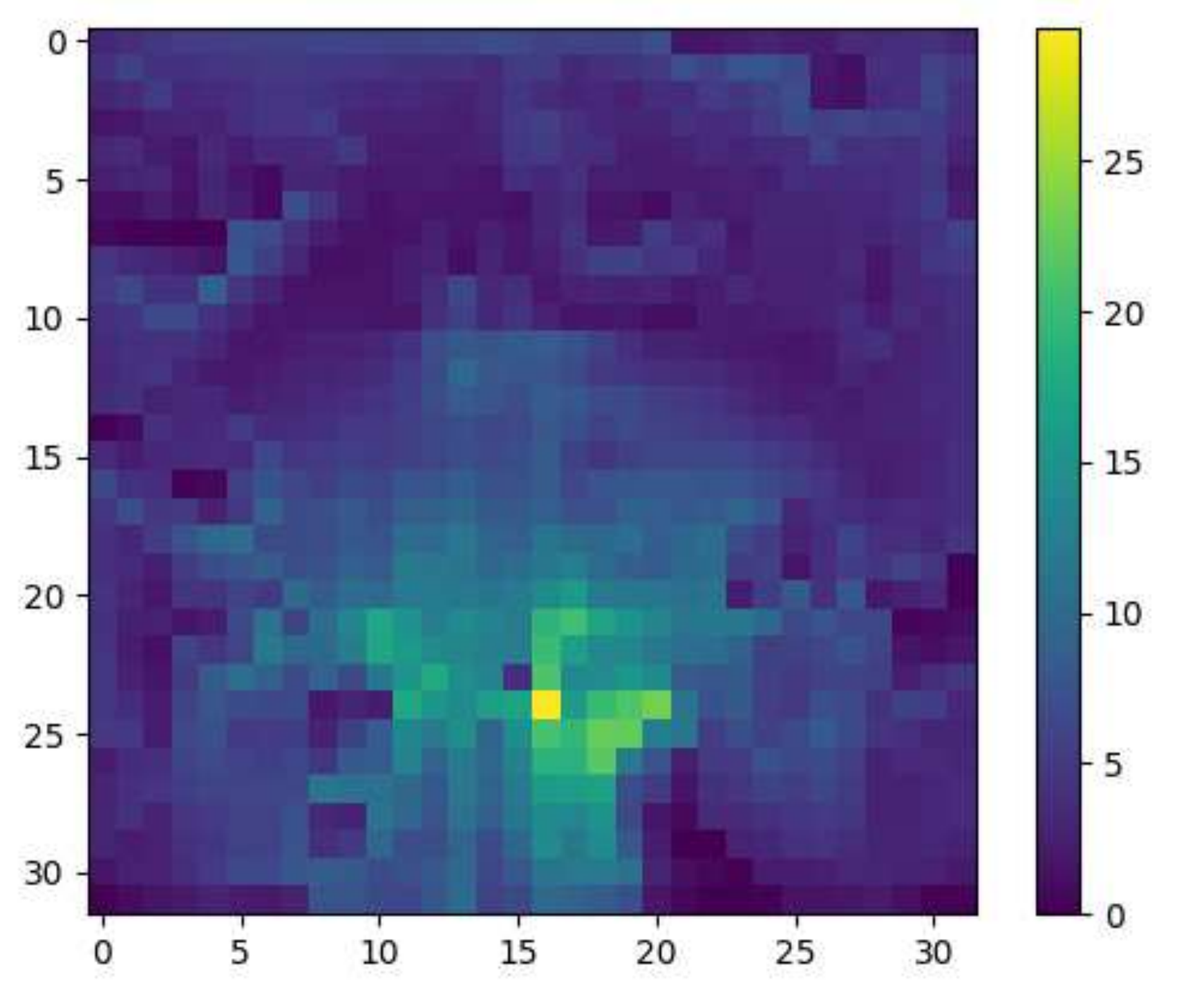}
\end{minipage}
\begin{minipage}[c]{0.24\textwidth}
\centering
\includegraphics[width=1\linewidth]{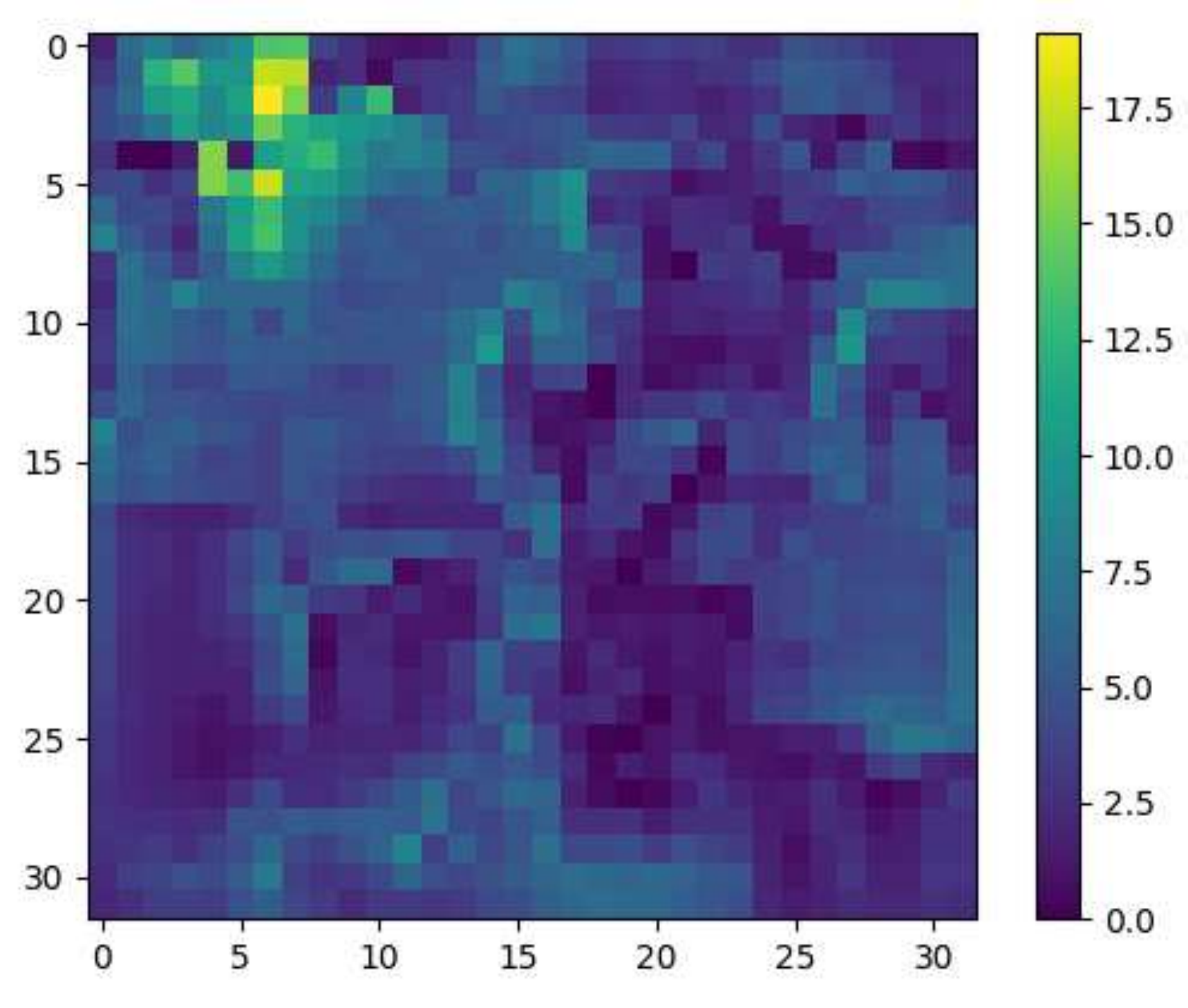}
\end{minipage}}
\subfigure[Value functions estimated by \emph{B1-Net}]{
\begin{minipage}[c]{0.24\textwidth}
\centering
\includegraphics[width=1\linewidth]{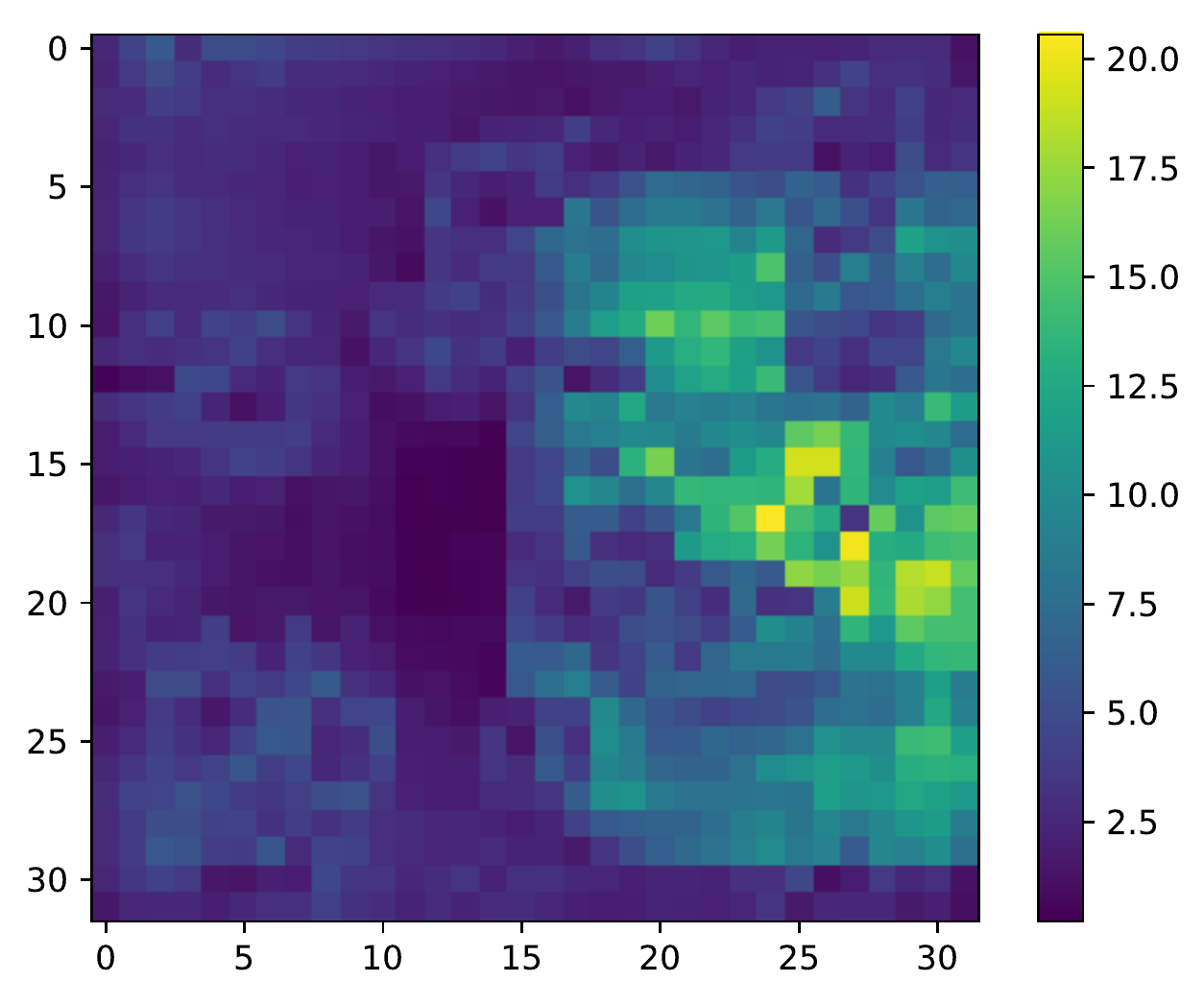}
\end{minipage}
\begin{minipage}[c]{0.24\textwidth}
\centering
\includegraphics[width=1\linewidth]{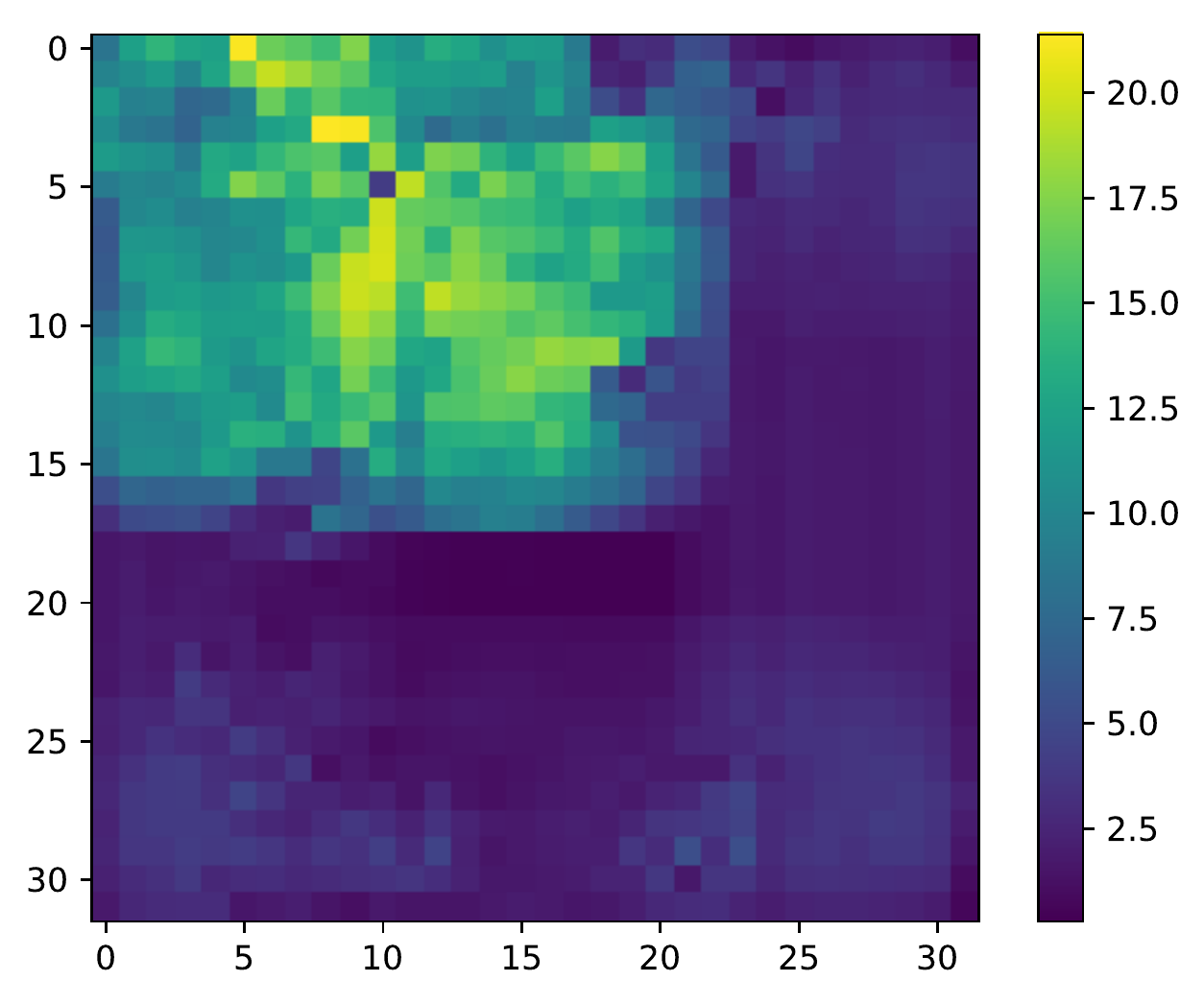}
\end{minipage}
\begin{minipage}[c]{0.24\textwidth}
\centering
\includegraphics[width=1\linewidth]{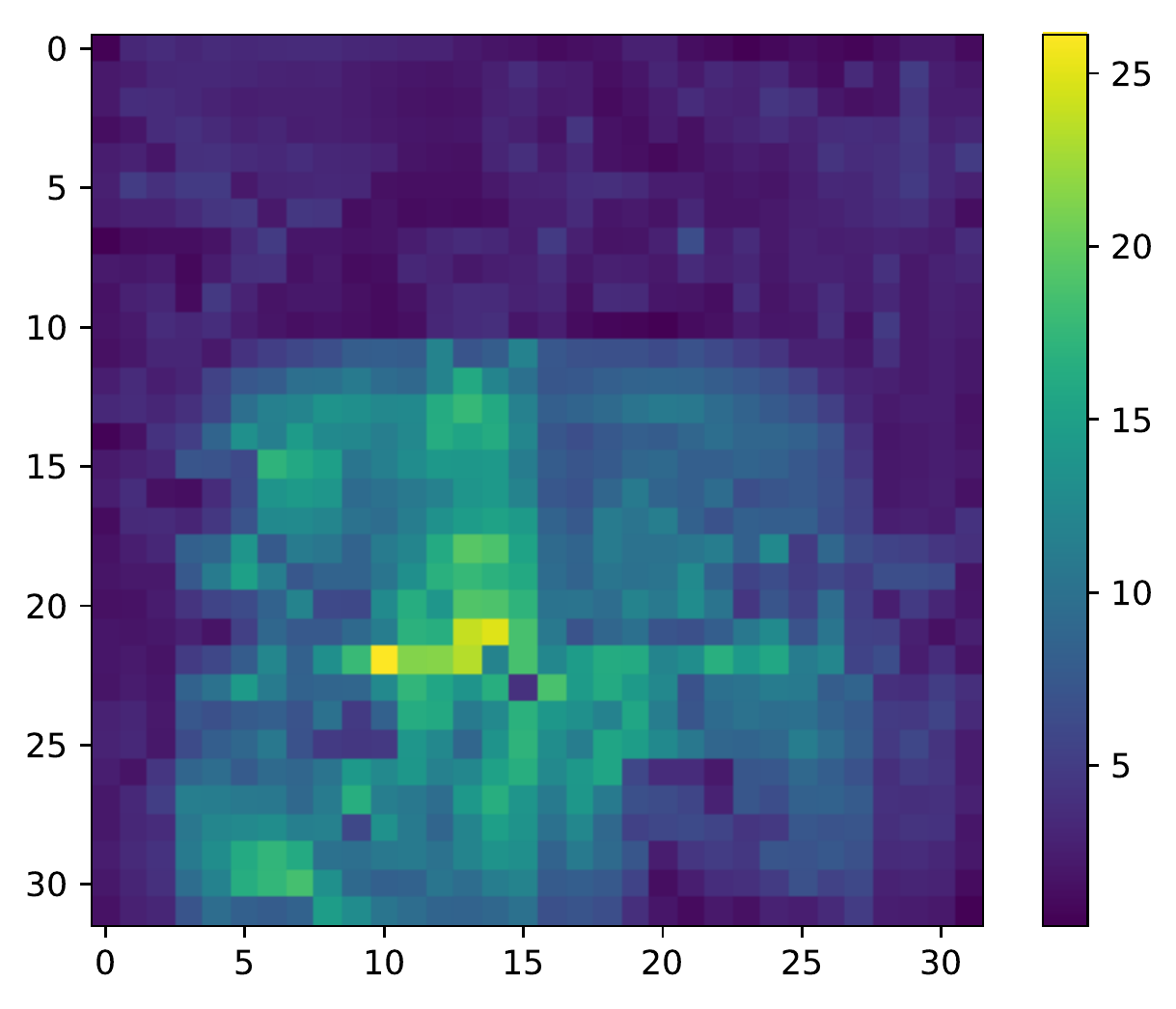}
\end{minipage}
\begin{minipage}[c]{0.24\textwidth}
\centering
\includegraphics[width=1\linewidth]{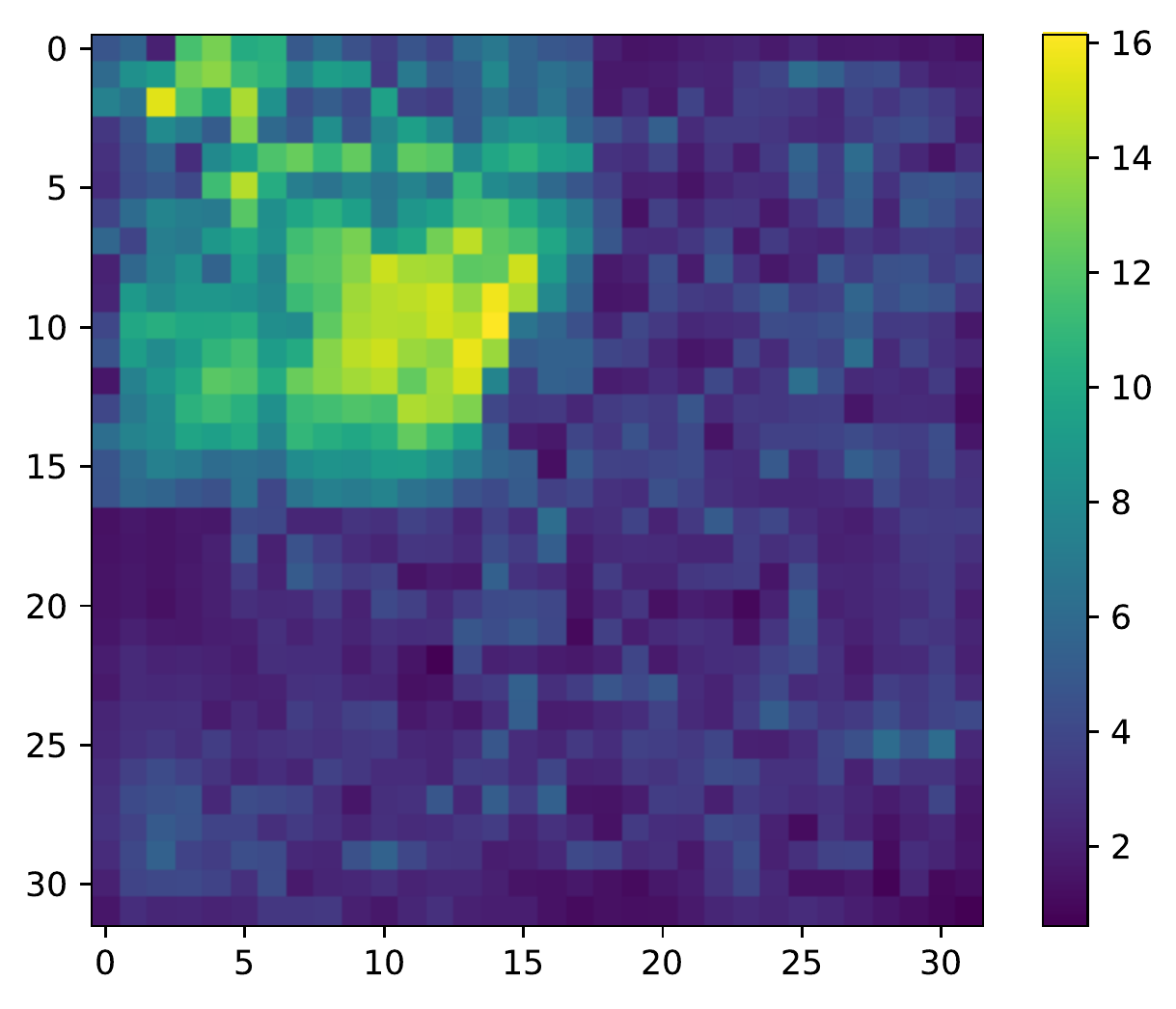}
\end{minipage}}
\subfigure[Value functions estimated by \emph{VIN}]{
\begin{minipage}[c]{0.24\textwidth}
\centering
\includegraphics[width=1\linewidth]{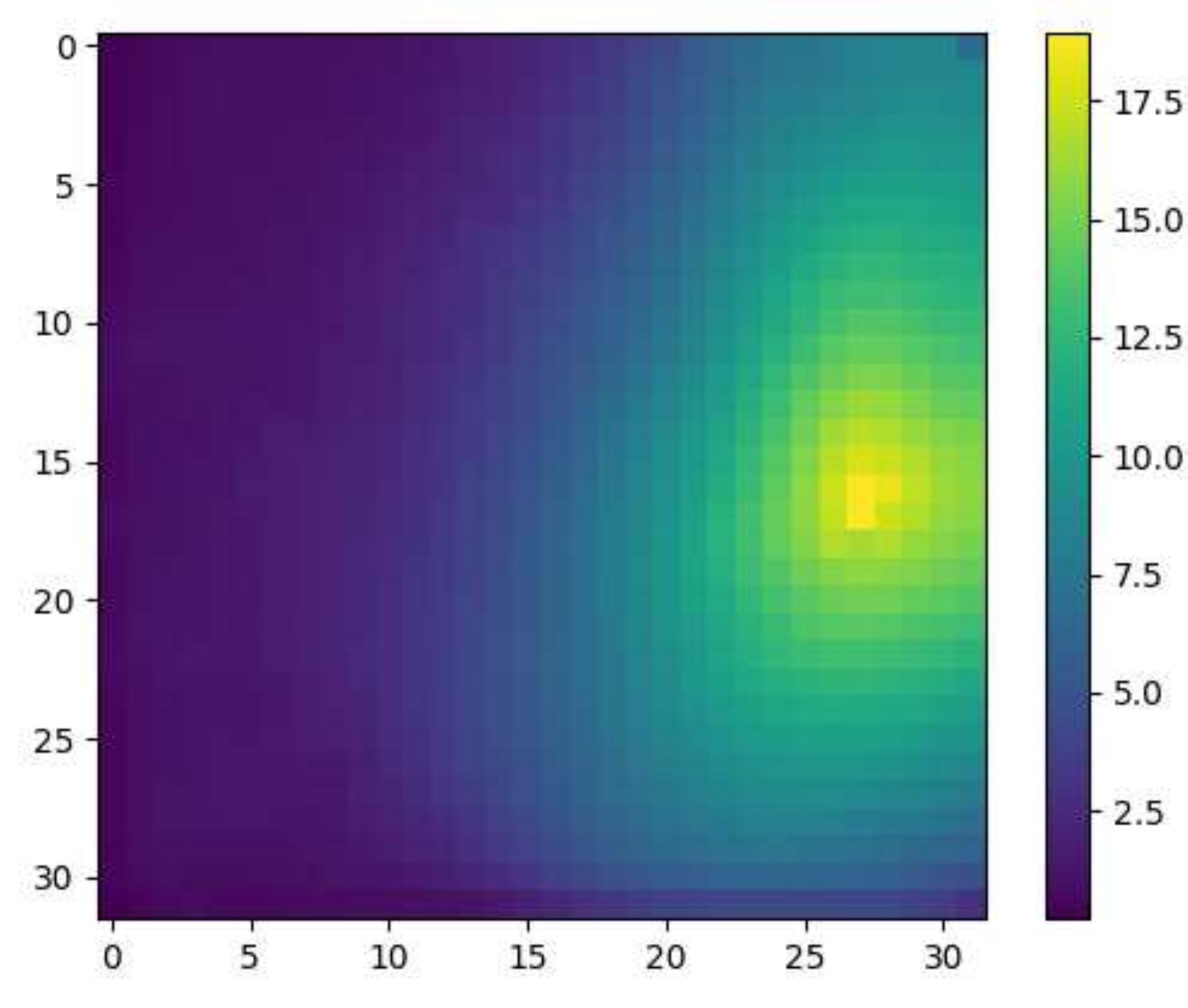}
\end{minipage}
\begin{minipage}[c]{0.24\textwidth}
\centering
\includegraphics[width=1\linewidth]{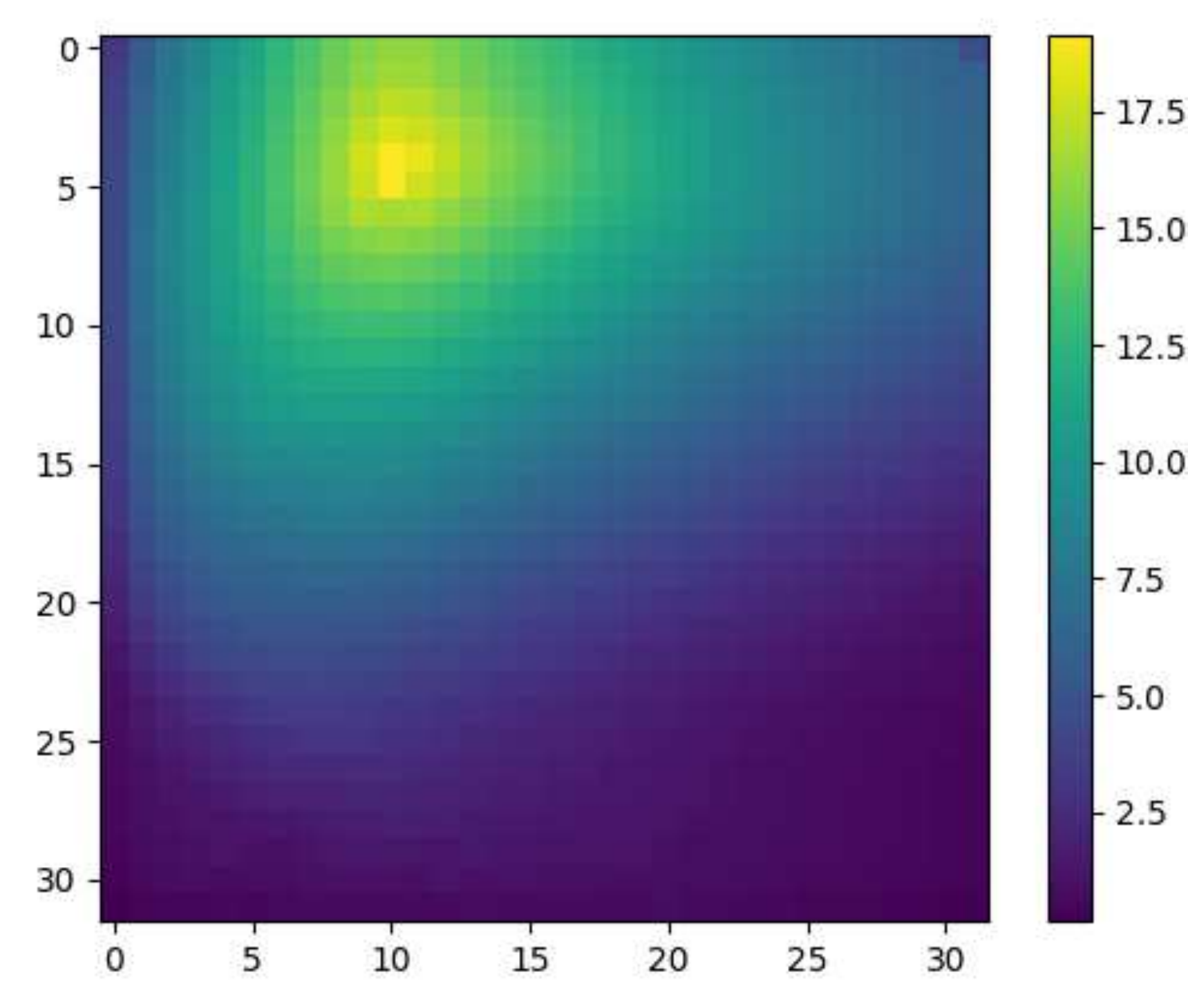}
\end{minipage}
\begin{minipage}[c]{0.24\textwidth}
\centering
\includegraphics[width=1\linewidth]{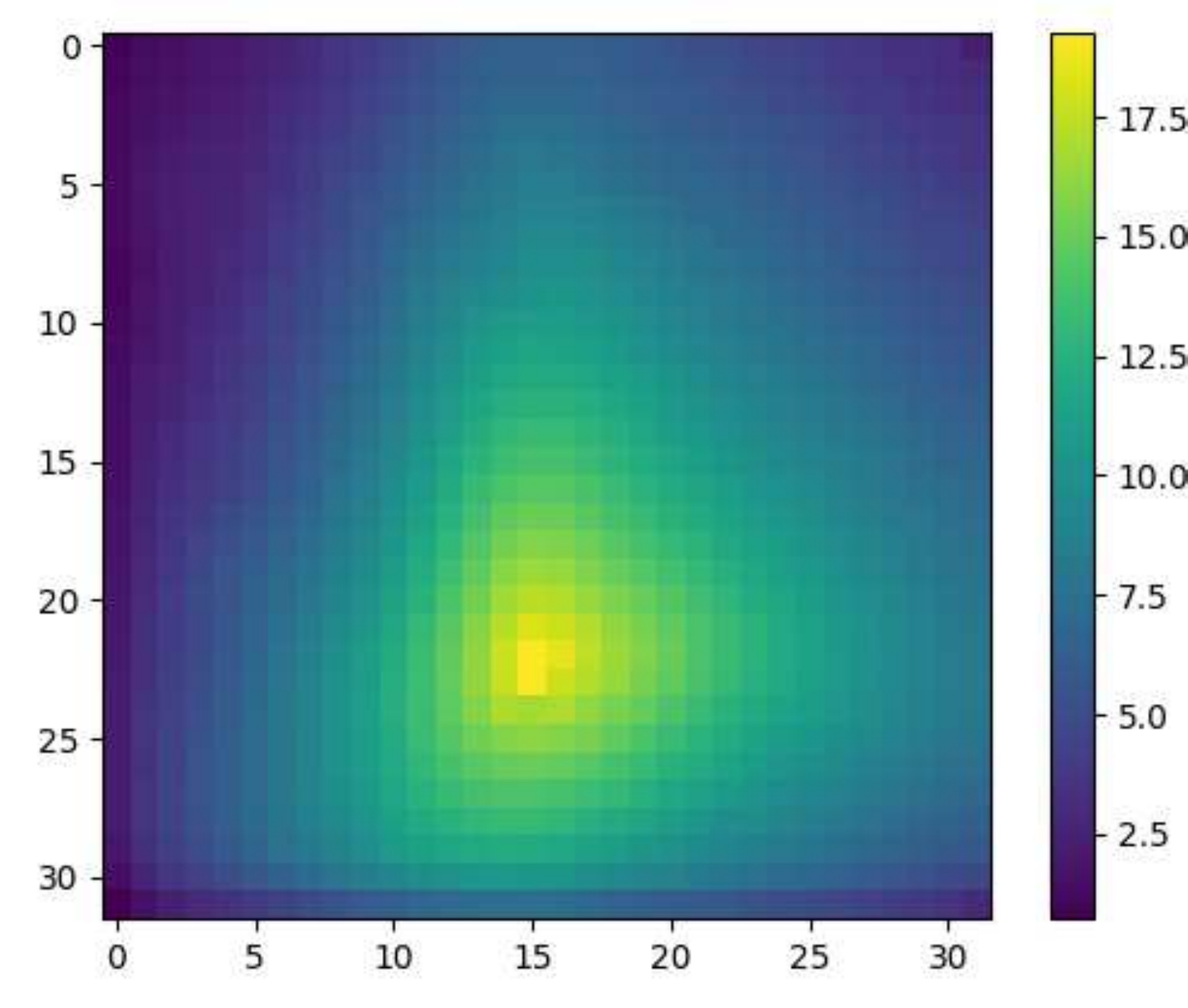}
\end{minipage}
\begin{minipage}[c]{0.24\textwidth}
\centering
\includegraphics[width=1\linewidth]{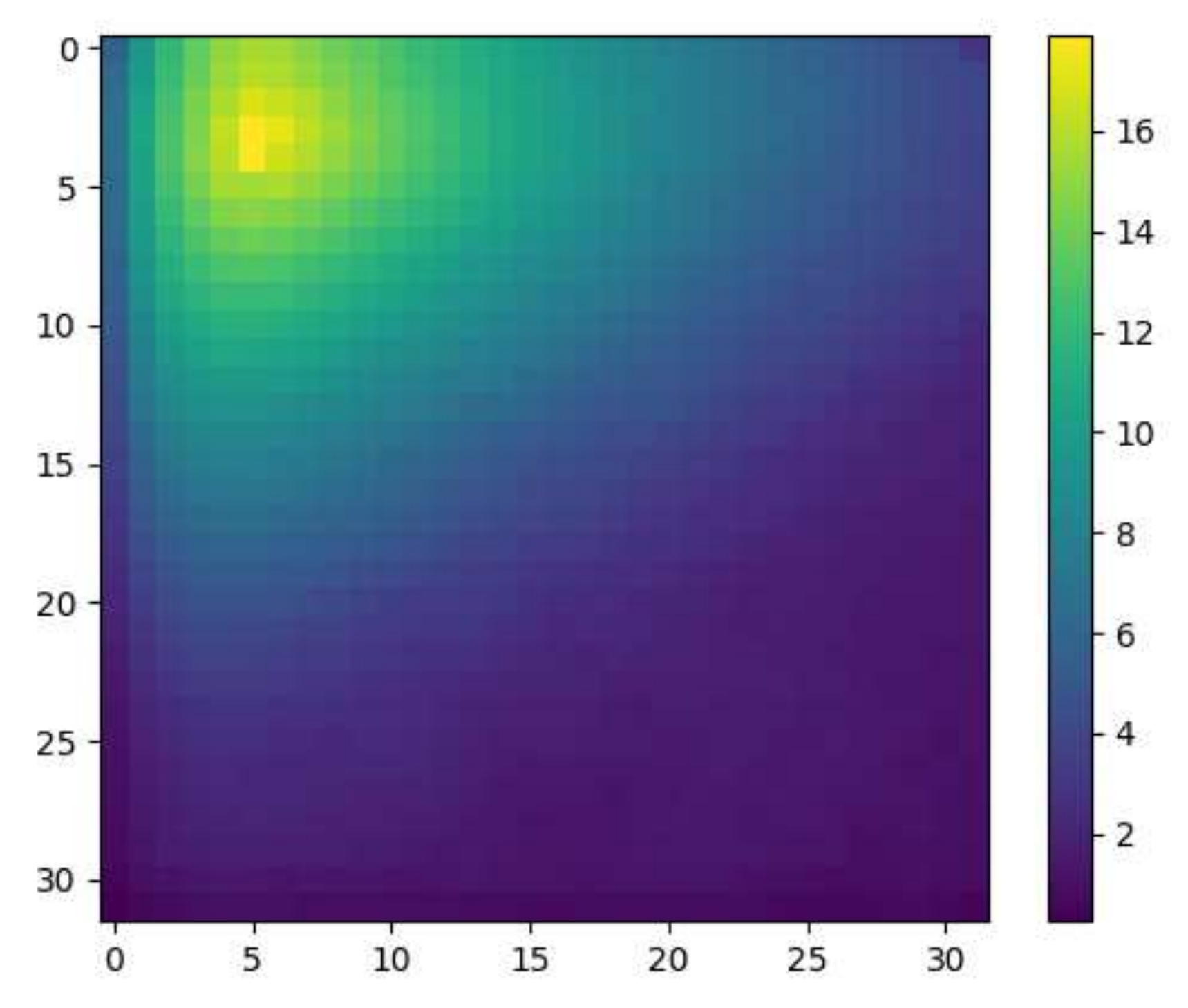}
\end{minipage}}
\caption{Visualization of value function layers.}
\label{Fig.4.6}
\end{figure*}

In this subsection, to test whether \emph{DB-Net} could keep its performance after ablating some of its components, model ablation experiments are conducted. Define  \emph{DB-Net} without branch one as \emph{B1-Net}. Then, derive \emph{B2-net} by replacing residual convolutional layers of \emph{B1-Net} with normal convolutional layers. As illustrated in Fig.~\ref{Fig.4.3} and TABLE~\ref{tab4.3}, without global deep features, the navigation accuracy and success rate of \emph{B1-Net} drop promptly compared with \emph{DB-Net}. Moreover, with only normal convolutional layers, training cost and error of \emph{B2-Net} remain at high levels, unable to provide reliable navigation policy for the Mars rover. Therefore, both of the two-branch architecture and the residual convolutional layers make indispensable contributions to the final performance of \emph{DB-Net}.
\begin{table}[H]
\centering
\caption{Results of model ablation experiments}\label{tab4.3}
\begin{tabular}{|c|c|c|c|c|c|c|}
\hline
28x28 grid map                    & \emph{DB-Net}     & \emph{B1-Net}     & \emph{B2-Net}       \\\hline
Training accuracy                 & \textbf{96.4\%}   & 86.2\%            & 13.8\%              \\\hline
Testing accuracy                  & \textbf{95.6\%}   & 85.8\%            & 12.8\%              \\\hline
Training success rate             & \textbf{96.0\%}   & 63.2\%            & 1.1\%              \\\hline
Testing success rate              & \textbf{93.3\%}   & 63.2\%            & 1.3\%              \\\hline
\end{tabular}
\end{table}
Moreover, to explore the inner mechanism of \emph{DB-Net}, the final value function layers ($\Pi_{\theta}$) of \emph{DB-Net}, \emph{B1-Net} and \emph{VIN} are contrasted in a visualized way. The value function layers estimates the action value distribution of current Martian images and target point. After being visualized, locations $(x_{1t},x_{2t})$ close to target point should be lighter (larger value) while location far from target point or near risky areas should be darker (smaller value). As demonstrated in Fig.~\ref{Fig.4.6}, the value functions estimated by by \emph{DB-Net} are more in coincidence with the original Martian images compared with \emph{B2-Net}. It is clear that risky areas are darker and the lighter locations are around target points in value function layers generated by \emph{DB-Net} from Fig.~\ref{Fig.4.6}. By contrast, \emph{B1-Net} without global deep features cannot estimate the value function as precisely as \emph{DB-Net}. \emph{VIN} also fails to recognize risky areas of Martian images evidently. Therefore, \emph{DB-Net} indeed has a remarkable capability of representing deep features and estimating the value distribution of current Martian environment.
\section{Conclusions}
In this paper, a novel deep neural network architecture---\emph{DB-Net} with double branches and non-recurrent structure is designed for dealing with Martian visual navigation problem. \emph{DB-Net} is able to determine the optimal navigation policy to target point directly from original Martian environment images without any prior knowledge. Moreover, compared with the existing best architecture---\emph{VIN}, \emph{DB-Net} achieves higher precision and efficiency. Most significantly, the average training time of \emph{DB-Net} is reduced by 45.8\%. In future research, more effective deep neural network architecture will be explored and the robustness of the architecture will be researched further.
\section{Acknowledgement}
This work was supported by the National Key Research and Development Program of China under Grant 2018YFB1003700, the Beijing Natural Science Foundation under Grant 4161001, the National Natural Science Foundation Projects of International Cooperation and Exchanges under Grant 61720106010, and by the Foundation for Innovative Research Groups of the National Natural Science Foundation of China under Grant 61621063.


\begin{thebibliography}{99}
\bibitem{ref1}Braun, R., Manning, R.: `Mars exploration entry, descent and landing challenges', Journal of Spacecraft Rockets, 2007, 44 (2), pp.310-323.

\bibitem{ref2}Matthies, L., Maimone, M., Johnson, A., et al.: `Computer Vision on Mars', Internal Journal of Computer Vision, 2007, 75 (1), pp.67-92.

\bibitem{ref3}Joseph, C., Arturo, R., Dave, F.: `Global path planning on board the Mars exploration rovers', Aerospace Conference, 2007, pp.1-11.
\bibitem{ref4}Sakuta, M., Takanashi, S., and Kubota, T.: `An image based path planning scheme for exploration rover', IEEE International Conference on Robotics and Biomimetics, 2011, pp.385-388.
\bibitem{ref5}Guo, Q., Zhang, Z., Xu, Y.: `Path-planning of automated guided vehicle based on improved Dijkstra algorithm', Control and Decision Conference, 2017, pp.7138-7143.
\bibitem{ref6}Chiang, C.H., Chiang, PJ., Fei, C.C., et al.: `A comparative study of implementing Fast Marching Method and A* search for mobile robot path planning in grid environment: Effect of map resolution', IEEE Workshop on Advanced Robotics and Its Social Impacts, 2007, pp.1-6.
\bibitem{ref7}Jeddisaravi, K., Alitappeh, R.J., Guimaraes, F.G.: `Multi-objective mobile robot path planning based on A* search', International Conference on Computer and Knowledge Engineering, 2017, pp.7-12.
\bibitem{ref8}Ferguson, D., Stentz, A.: `Using interpolation to improve path planning: The Field D* algorithm', Journal of Field Robotics, 2006, 23 (2), pp.79-101.
\bibitem{ref9}Shi, J., Liu, C., Xi, H.: `A framed-quadtree based on reversed D* path planning approach for intelligent mobile Robot ', Journal of Computers, 2012, 7 (2), pp.464-469.

\bibitem{ref10}Wooden, D.T.:`Graph-based Path Planning for Mobile Robots', thesis, Georgia Institute of Technology, 2006

\bibitem{ref11}Bassil Y.: `Neural network model for path-planning of robotic rover systems', International Journal of Science and Technology, 2012, 2 (2), pp.94-100.
\bibitem{ref12}Zeng, C., Zhang, Q., Wei, X.: `Robotic global path-planning based modified genetic algorithm and A* algorithm', International Conference on Measuring Technology and Mechatronics Automation, 2011, pp.167-170.
\bibitem{ref13}Kang, HI., Lee, B., Kim, K.: `Path planning algorithm using the particle swarm optimization and the improved Dijkstra algorithm', Workshop on Computational Intelligence and Industrial Application, 2009, 17 (4), pp.1002-1004.
\bibitem{ref14}Gu, J., Wang, Z., Kuen, J., et al.: `Recent advances in convolutional neural networks ', arXiv preprint arXiv:1512.07108, 2015.
\bibitem{ref15}Krizhevsky, A., Sutskever, I., Hinton, G., E.: `Imagenet classification with deep convolutional neural networks', Advances in neural information processing systems, 2012, pp.1097-1105.
\bibitem{ref16}Huang, J., Guadarrama, S., Murphy, K., et al.: `Speed/accuracy trade-offs for modern convolutional object detectors', arXiv preprint arXiv:1611.10012, 2016.
\bibitem{ref17}Zhu, Y., Mottaghi, R., Kolve, E., et al.: `Target-driven visual navigation in indoor scenes using deep reinforcement learning', Proceedings of the International Conference on Robotics and Automation, 2017, pp.3357-3364.
\bibitem{ref18}Levine, S., Finn, C., Darrell, T., et al.: `End-to-end training of deep visuomotor policies', Journal of Machine Learning Research, 2015, 17 (1), pp.1334-1373.
\bibitem{ref19}Tanner, C., Roberto, F., Richard, L., et al.: `A deep learning approach for optical autonomous planetary relative terrain navigation', AAS/AIAA Spaceflight Mechanics Meeting, 2017, pp.329-338.
\bibitem{ref20}Maturana, D., Scherer, S.: `3D convolutional neural networks for landing zone detection from LiDAR', International Conference on Robotics and Automation, 2015, pp.3471-3478.
\bibitem{ref21}Tamar, A., Wu, Y., Thomas, G., et al.: `Value iteration networks', In Advances in Neural Information Processing Systems, 2016, pp.2146-2154.
\bibitem{ref22}Khan, A., Zhang, C., Atanasov, N., et al.: `Memory augmented control networks', arXiv preprint arXiv:1709.05706, 2017
\bibitem{ref23}Bellman, R.: `Dynamic programming', Princeton University Press, 1957.
\bibitem{ref24}Bertsekas, D.P., Bertsekas, D.P., Bertsekas, D.P., et al.: `Dynamic programming and optimal control', Athena Scientific, 4th edition, 2012.
\bibitem{ref25}Sutton, R.S., Barto, A.G.: `Reinforcement learning: An introduction', MIT Press, 1998.
\bibitem{ref26}Li, Y.: `Deep reinforcement learning: An overview', arXiv preprint arXiv:1701.07274, 2017.
\bibitem{ref27}Attia, A., Dayan, S.: `Global overview of Imitation Learning', arXiv preprint arXiv:1801.06503, 2018.
\bibitem{ref28}Goodfellow, I., Bengio, Y., and Courville, A., et al: `Deep Learning', MIT Press, 2016.
\bibitem{ref29}Harris,D., Harris, S.: `Bergstrom, W.J., et al.: `Digital design and computer architecture', Chian Machine Press, 2014.
\bibitem{ref30}He, K., Zhang, X., Ren, S., et al.: `Deep residual learning for image recognition ', IEEE Conference on Computer Vision and Pattern Recognition, 2016, pp.770-778.
\bibitem{ref31}McEwen, S.A., Eliason, M.E., Bergstrom, W.J., et al.: `Mars Reconnaissance Orbiter's High Resolution Imaging Science Experiment (HiRISE)', Journal of Geophysical Research Planets, 2007, 112(E05S02), pp.1-40.
\bibitem{ref32}Canny, J.: `A Computational Approach To Edge Detection', IEEE Transaction on Pattern Analysis and Machine Intelligence, 1986, 8(6), pp.679-698.


\end{thebibliography}
\end{document}